\documentclass[12pt]{article}
\usepackage{amsmath}
\usepackage{mathtools}
\usepackage{amssymb} 
\usepackage{amsthm}
\usepackage{bbm}
\usepackage{graphicx,epsf}
\usepackage{epstopdf}
\usepackage{enumerate}
\usepackage{caption}
\usepackage{natbib}
\usepackage{bm}
\usepackage{color}
\usepackage{fourier}
\usepackage{floatrow}

\graphicspath{{figures/}} 

\newtheorem{remark}{Remark}
\newtheorem{theorem}{Theorem}
\newtheorem{proposition}{Proposition}
\newtheorem{definition}{Definition}

\DeclareMathOperator*{\var}{\mbox{Var}}
\DeclareMathOperator*{\pen}{\mathcal{P}} 

\DeclareMathOperator{\WCSS}{\mbox{WCSS}}
\DeclareMathOperator{\Exp}{\mathbb{E}}

\newfloatcommand{capbtabbox}{table}[][\FBwidth]

\newcommand{\by}{\bm{y}}
\newcommand{\bx}{\bm{x}}
\newcommand{\bmu}{{\bm{\mu}}}
\newcommand{\bX}{\bm{X}}

\author{Jakob Raymaekers$^{1}$ \and Ruben H. Zamar$^2$}
\date{%
    $^1$ KU Leuven, Department of Mathematics,\\
		Celestijnenlaan 200B, 3001 Leuven, Belgium.\\%
    $^2$ University of British Columbia, Department of Statistics,\\
		Earth Sciences Building, 2207 Main Mall, Vancouver, Canada.\\[2ex]%
    \today
}

\begin{document}
\title{\Large Regularized $K$-means through hard-thresholding}
\maketitle

\begin{abstract}
We study a framework of regularized $K$-means methods based on direct penalization of the size of the cluster centers. Different penalization strategies are considered and compared through simulation and theoretical analysis. Based on the results, we propose HT $K$-means, which uses an $\ell_0$ penalty to induce sparsity in the variables. Different techniques for selecting the tuning parameter are discussed and compared. The proposed method stacks up favorably with the most popular regularized $K$-means methods in an extensive simulation study. Finally, HT $K$-means is applied to several real data examples. Graphical displays are presented and used in these examples to gain more insight into the datasets. 
\end{abstract}

\clearpage
\section{Introduction}

Clustering is one of the most commonly used unsupervised learning techniques. The goal of clustering is to partition the data into homogeneous groups. We focus on $K$-means, a method introduced by \cite{Steinhaus1956} and popularized by \cite{macqueen1967some}. We assume that we observe a $n \times p$ data matrix $\bX$, containing $n$ observations $\bx_1, \ldots, \bx_n$ in $p$ dimensions. The $K$-means clustering algorithm tries to find the $K$ cluster centers $\bmu_1,  \ldots, \bmu_K$ that minimize the within-cluster sum of squares (WCSS) defined as
\begin{equation}\label{eq:kmeans}
\WCSS = \frac{1}{n}\sum_{i = 1}^{n}{\min_{k \in \{1,\ldots, K\}}{||\bx_i - \bmu_{k}||_2^2}}.
\end{equation} 
Based on these centers, the data can be partitioned into $K$ clusters by assigning each observation to the cluster corresponding to the nearest (in Euclidean distance) cluster center. Despite being over 50 years old, the $K$-means algorithm is still very popular and widely used in a variety of scientific fields, see \cite{jain2010data} for a recent overview.\par
\begin{figure}
\begin{floatrow}
\ffigbox{%
  \includegraphics[width = 0.5\textwidth]{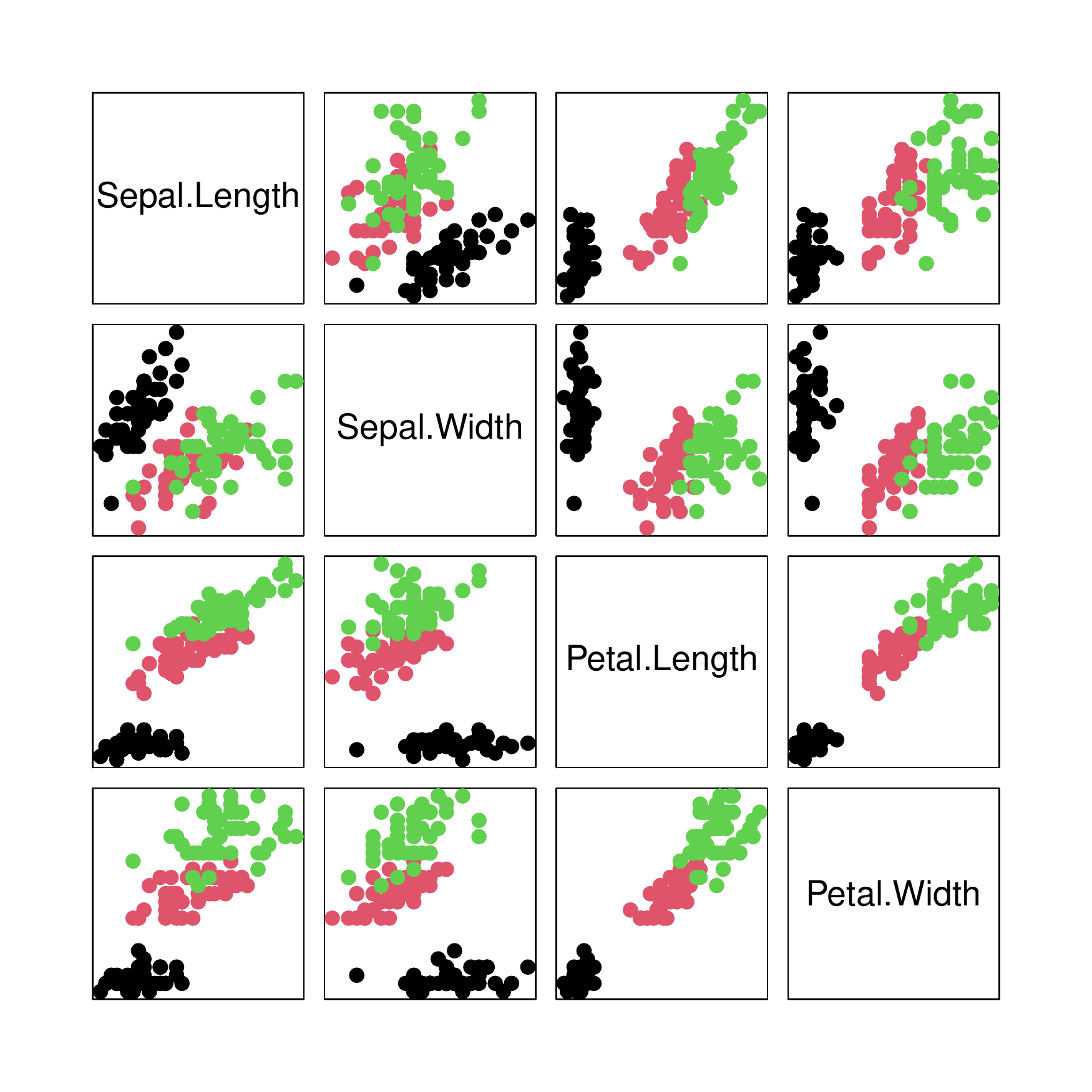}
}{%
  \caption{Pairs plot of the Iris data}%
	\label{fig:iris}
}
\capbtabbox[5cm]{%
 \fontsize{12}{12}\selectfont
 \begin{tabular}{ll}
  \hline
Used variables & ARI \\ 
  \hline
(1) & 0.38 \\ 
  (2) & 0.15 \\ 
  (3) & 0.85 \\ 
  (4) & \textbf{0.89} \\ 
  (1, 2) & 0.6 \\ 
  (1, 3) & 0.7 \\ 
  (1, 4) & 0.57 \\ 
  (2, 3) & 0.8 \\ 
  (2, 4) & 0.8 \\ 
  (3, 4) & \textbf{0.89} \\ 
  (1, 2, 3) & 0.7 \\ 
  (1, 2, 4) & 0.61 \\ 
  (1, 3, 4) & 0.73 \\ 
  (2, 3, 4) & 0.87 \\ 
  (1, 2, 3, 4) & 0.73 \\ 
   \hline
\end{tabular}
}{%
  \caption{ARI of clustering the Iris data using subsets of variables. The best performance is reached by clustering based on variable 4 or the combination of variable 3 and 4.} 
\label{tab:iris}
}
\end{floatrow}
\end{figure}

Whereas in classical $K$-means, all $p$ features are used to partition the data, it might be desirable to identify a subset of features that partitions the data particularly well. This feature selection may lead to a more interpretable partitioning of the data and more accurate recovery of the ``true'' clusters. We notice that feature selection is not only relevant for scenarios where $p >> n$, but also when $p < n$. The former scenario, with (many) more variables than observations, is likely to include many uninformative variables which do not contribute to clustering the data and are better left out of the analysis. The latter scenario is typically easier to work with, but may also produce datasets with variables which do not contribute to and rather difficult the partitioning of the data. To illustrate this, we consider the classical example of Fisher's Iris data \citep{Fisher1936}, collected by \cite{Anderson1935}. The data consists of 150 iris flowers which are described by 4 variables characterizing the dimensions of their sepal and petal. The flowers can be subdivided in 50 samples of each of three types of iris: Iris setosa, versicolor, and virginica. Figure \ref{fig:iris} shows a plot of the data in which the different iris types appear in different colors. From this plot it is clear that not all the variables separate the flowers equally well. This becomes more evident after we cluster this dataset using the $K$-means algorithm on all possible subsets of variables. Table \ref{tab:iris} shows the adjusted rand index (ARI) for each of these clusterings. The ARI measures the agreement between an estimated partition and the ``true'' partition. An ARI of 1 corresponds with perfect clustering. Interestingly, $K$-means performs best (ARI = 0.89) when variable 4 alone or variable 3 and 4 are used for the clustering. This ARI value is substantially higher than the ARI of 0.73 obtained when clustering the data with all 4 variables. This example illustrates that even for datasets with very few variables, feature selection can be very useful.

	%


When it comes to the $K$-means algorithm, an influential reference for the practice of combining feature selection with clustering is the paper by \cite{Witten2010} called \textit{sparse $K$-means}. In this approach, the $K$-means objective function of Equation \ref{eq:kmeans} is rewritten as a maximization problem, in which a vector of feature weights is introduced. An appropriate penalization strategy applied to the vector of feature weights induces sparsity in the variables and shrinkage in the estimated cluster centers. The new objective function can be optimized by iteratively maximizing it with respect to the cluster centers and the cluster memberships. \cite{Sun2012} proposed another regularized $K$-means approach based on direct penalization of the size of the cluster centers using a group-lasso penalty. The group-lasso penalty also induces sparsity in the features and shrinkage in the estimated cluster centers.\par
We further study the regularized $K$-means approach and build a general regularization framework through direct penalization of the size of the cluster centers. We consider several common penalization strategies including lasso, ridge, group-lasso and $\ell_0$-type penalties. We present a general iterative algorithm for the estimation of the cluster centers. The algorithm gives insight into the effect of the different penalties on the estimated cluster centers. A combination of theoretical analysis and numerical studies shows the advantage of the proposed hard-thresholding $K$-means algorithm. The rest of the paper is organized as follows.\par
Section \ref{sec:meth} introduces the proposed framework for regularized $K$-means clustering and the penalties under consideration. It also presents the algorithm for the implementation of the clustering method for the different penalties. Section \ref{sec:sim_pen} presents a simulation study aimed at the identification of the most promising penalty types within the framework. A theoretical analysis of the proposed method is presented in Section \ref{sec:theo}. Section \ref{sec:lambdasel} discusses the selection of the regularization parameter and \ref{sec:sim_comp} compares the proposed HT $K$-means algorithm with well known competitors. Finally, Section \ref{sec:examples} illustrates the method on a number of real data examples.

\clearpage
\section{Methodology}\label{sec:meth}
\subsection{Regularized $K$-means clustering}

Let $\bx_1, \ldots, \bx_n$ denote $n$ observations from a $p$-variate distribution which we want to cluster. Suppose that the variables are standardized, i.e. we have $\frac{1}{n}\sum_{i=1}^{n}{x_{i,j}} = 0$ and $\frac{1}{n}\sum_{i=1}^{n}{x_{i, j}^2} = 1$ for every variable $j$. We consider the following general form of a regularized $K$-means objective function. Given the number of clusters $K$, let $\bmu \in \mathbb{R}^{K \times p}$ be a $K \times p$ matrix of cluster centers and $C = \{C_1, \ldots, C_K\}$ a collection of $K$ disjoint sets of cluster indices satisfying $\bigcup\limits_{k=1}^{K} C_k = \{1, \ldots, n\}$. We look for $\widehat{\bmu}$ and $\widehat{C} = \{\widehat{C}_1, \ldots, \widehat{C}_K\}$  which minimize
\begin{equation}\label{eq:regkmeans}
\frac{1}{n}\sum_{k = 1}^{K}\left\{\sum_{i \in C_k}{||x_i - \textstyle \bmu_{k, \cdot}||_2^2}\right\} + \lambda \pen(\bmu),
\end{equation} 
where $\lambda \geq 0$ is a tuning parameter and $\pen(\bmu)$ is a penalty that depends on the cluster centers $\bmu$. The notation $A_{i,\cdot}$ and $A_{\cdot,j}$ indicates the $i$-th row and $j$-th column of a matrix $A$ respectively, and is used throughout the article. The first term in Equation \ref{eq:regkmeans} is the classical $K$-means objective (\ref{eq:kmeans}). Depending on how $\pen$ is defined, different optimization problems arise. The important point is that the penalization is done based on $\bmu$, which is not always the case in existing proposals for regularized $K$-means clustering such as that in \cite{Witten2010}.\par
The intuition for penalizing the size of the cluster centers stems from the fact that we expect that when a variable does not contribute to the partitioning of the data, its estimated cluster centers will be close to the overall mean of the variable (i.e. 0). The following asymptotic argument may add to the intuition. Consider the optimal asymptotic value of the classical $K$-means objective function of Equation \ref{eq:kmeans}: 
$\mbox{obj} = \int\min_{k \in \{1,\ldots, K\}}{||\bx - \bmu_{k}||_2^2} Q(d\bx)$, where $Q$ denotes the distribution of a $m$-variate random variable $X$ where $m <p$.  In this asymptotic setting each cluster corresponds to a region in $\mathbb{R}^m$. Let $R_1,\ldots, R_K$ be these regions. Now keep the cluster assignments fixed and suppose we add an additional variable which is uninformative, in the sense that it is completely independent of the existing clustering.

Denote the new $(m+1)$-dimensional random vector by $X^*$ and its distribution by $Q^*$. Let $\mbox{obj}$ and $\mbox{obj}^*$ be the old and new values of the objective function. Then we have that 
$$ \mbox{obj} + 1 \geq \mbox{obj}^* \geq \mbox{obj} + \sum_{k=1}^K{\min_{a\in \mathbb{R}}{\int{(x_{m+1} - a)^2 I_{x_{1:m} \in R_k} Q^*(dx)}}},$$
where $x_{1:m}$ and $x_{m+1}$ denote the first $m$ and the last element of the vector $x$ respectively. Now due to independence, we have $$\min_{a\in \mathbb{R}}{\int{(x_{m+1} - a)^2  I_{x_{1:m} \in R_k} Q^*(dx)}} = \Pr(X \in R_k) \min_{a\in \mathbb{R}}{\int{(y - a)^2 Q_{m+1}^*(dy)}}$$ for all $k = 1,\ldots, K$, where $Q_{m+1}$ denotes the marginal distribution of the added variable. We thus find that $a$ should be equal to the mean of the added variable, independent of $k$. As the variables are standardized, we find that the cluster center of the added variable should be at 0 for every cluster $k = 1,\ldots, K$ and that $\mbox{obj}^* = 1 + \mbox{obj}$. Of course, this is a simplified argument as it assumes that the cluster assignments do not change when adding the extra variable. In reality, the asymptotic assignments may change if the added variable dominates the clustering structure but this is rather unlikely under the assumption that at least a few informative variables are present and that the variables are standardized.\\

\noindent Throughout the paper we will consider several options for the penalty type which we name after their familiar counterparts from regularized regression:

\begin{align*}
&\mbox{\textbf{best-subset}: } &\textstyle \pen_0(\bmu) = \sum_{j=1}^{p}{I_{||\textstyle \bmu_{\cdot,j}||_2 > 0}}\\
&\mbox{\textbf{lasso}: } &\textstyle \pen_1(\bmu) = \sum_{j=1}^{p}{||\textstyle \bmu_{\cdot,j}||_1}\\
&\mbox{\textbf{ridge}: } &\textstyle \pen_2(\bmu) = \sum_{j=1}^{p}{||\textstyle \bmu_{\cdot,j}||_2^2}\\
&\mbox{\textbf{group-lasso}: } &\textstyle \pen_3(\bmu) = \sum_{j=1}^{p}{||\textstyle \bmu_{\cdot,j}||_2}\\
\end{align*}

\noindent The penalty on $\bmu$ balances the size of the cluster centers and their contribution to the objective function. Essentially, it implies that the cluster centers can be large only if they reduce the WCSS sufficiently. When a certain variable has only zero cluster centers, this variable becomes redundant in the clustering. An algorithm to optimize (\ref{eq:regkmeans}) is derived in the next section. This algorithm also helps to better understand the effect of the different penalties on the clustering results.\\

\subsection{Computation}

\noindent In order to compute the cluster centers and indices resulting from the optimization in Equation \ref{eq:regkmeans}, we use an adaptation of Lloyd's algorithm \citep{lloyd1982least} for classical $K$-means:\\

Given an initial set of cluster centers:
\begin{enumerate}
\item Update the cluster indices $\hat{C}$ by minimizing Equation \ref{eq:regkmeans} with respect to the cluster indices while keeping the cluster centers fixed.
\item Update the cluster centers $\hat{\bmu}$ by minimizing Equation \ref{eq:regkmeans} with respect to the cluster centers while keeping the cluster indices fixed.
\item Repeat 1. and 2. until convergence.\\
\end{enumerate}

\noindent It is clear that in step 1., each point is assigned to the cluster corresponding with the nearest cluster center (in Euclidean distance), since keeping the cluster centers fixed also implies that the penalty term of Equation \ref{eq:regkmeans} is fixed. This is similar to the classical $K$-means objective function and the corresponding Lloyd's algorithm \citep{lloyd1982least}. Step 2 minimizes the objective function with respect to the cluster centers while keeping the cluster indices fixed. The penalty parameter is now dependent on the cluster centers $\hat{\bmu}$, and the resulting updated centers are therefore not equal to the cluster means as is the case for classical $K$-means. The following proposition presents the updating equations for the penalties under consideration. The proof can be found in the Supplementary Material.

\begin{proposition}\label{prop:update}
Suppose that we have an assignment of the elements into $K$ clusters $C_1, \ldots, C_K$. Let $|C_k|$ be the number of elements in cluster $k$. Denote with $\bmu^*$ the $K \times p$ matrix of cluster means and with $M\in \{0,1\}^{n\times K}$ the matrix for which $M_{i,k} = 1$ if $\bx_i$ is in cluster $k$. Keeping this assignment fixed, minimizing the objective function in Equation \ref{eq:regkmeans} with respect to the $K \times p$ matrix of cluster centers $\bmu$ yields:

\begin{align*}
&\textstyle \pen(\bmu) = \pen_0(\bmu) \mbox{  yields }& 
\textstyle \bmu_{k, j}& = \begin{cases}
\bmu_{k, j}^* \mbox{ if } ||X||_2^2 > ||X - M\bmu_{\cdot, j}^*||_2^2 + n\lambda\\
0 \mbox{ else } 
\end{cases}
\\
&\textstyle \pen(\bmu) =\pen_1(\bmu)  \mbox{  yields }&\textstyle \bmu_{k, j}& = \max\left(0, 1 - \frac{n \lambda}{2 |C_k|\left|\bmu_{k, j}^*\right|}\right) \textstyle \bmu_{k, j}^*\\
&\textstyle \pen(\bmu) =\pen_2(\bmu)  \mbox{  yields }&\textstyle \bmu_{k, j}& = \frac{1}{1+\frac{n \lambda}{|C_k|}} \textstyle \bmu_{k, j}^* \\
&\textstyle \pen(\bmu) =\pen_3(\bmu)  \mbox{  yields }& \textstyle \bmu_{k, j}& =
 \frac{1}{1 + \frac{n \lambda}{ (2 |C_k| ||\bmu_{\cdot, j}||_2)}} \textstyle \bmu_{k, j}^* \mbox{ if } \bmu_{\cdot, j}\neq \bm 0
\\
\end{align*}
\end{proposition}

\noindent These updating equations provide additional insight into the effect of the different penalty types. $\pen_0$ leads to hard thresholding. It is the literal translation of ``include a variable in the clustering if it sufficiently reduces the WCSS''. If the variable is included (i.e. the corresponding vector of cluster centers is non-zero), the cluster centers are given by the means within each cluster as in classical $K$-means. $\pen_1$ is a lasso-type penalty. It shrinks some of the coefficients to exactly zero, and others are translated towards 0. The updating equation uses a soft-thresholding operator, and bears strong resemblance to solution of lasso regression with orthonormal covariates. $\pen_2$ is a ridge-type penalty and shrinks all the cluster centers towards zero without setting them to zero exactly. Like in regression, it does not induce any sparsity and the shrinkage is proportional to $1/\lambda$. $\pen_3$ is the only penalty which does not have an explicit updating equation, as the right hand side contains the euclidean norm of the vector of centers $||\bmu_{\cdot, j}||_2$. The solution is thus implicit and can be found through an iterative algorithm. This penalty induces sparsity in the cluster centers, while shrinking in a ridge-type fashion within each center that is not shrunk to zero.\\

\begin{remark}[Size-dependent penalties]
Note that the cluster sizes play a role in the update steps of penalties $\pen_1$, $\pen_2$ and $\pen_3$. These seem to be somewhat unnatural and can be removed by including penalties which depend linearly on the size of the clusters. For example, if we replace $\lambda$ by $\lambda_i = \lambda \frac{|C_i|}{n}$, we would obtain more elegant expressions as both $n$ and $|C_i|$ would disappear in the updating equations. For model-based clustering, this was done by \cite{bhattacharya2014}. We did not pursue this path any further since it did not yield substantial improvements for $\pen_1$ and $\pen_2$ in the simulation study and makes the optimization slightly slower. Especially for $\pen_3$ it is not immediately clear how this should be implemented without a substantial increase in computational cost. We suspect it may have potential when the true cluster sizes are very unbalanced. Note that this does not affect the $\pen_0$ penalty.
\end{remark}

\begin{remark}[Adaptive penalties]
In addition to making the penalties dependent on the cluster sizes, there is the option to make them adaptive. This idea was introduced by \cite{zou2006} in the context of lasso regression to obtain both $\sqrt{n}$-consistency as well as consistent variable selection. It was also used by \cite{Sun2012} in their version of regularized $K$-means clustering. It can be implemented by replacing $\lambda$ in the updating equations of proposition \ref{prop:update} by $\lambda_j = \frac{\lambda}{||\bmu_{\cdot, j}^*||_2}$.
\end{remark}

\noindent Like the classical $K$-means problem, the regularized version is NP-hard \citep{dasgupta2008, aloise2009} and Lloyd's algorithm yields only locally optimal solutions. Therefore, the $K$-means algorithm is typically run using several starting values, after which the solution yielding the lowest objective function is retained. For the regularized $K$-means problem, one could take the starting centers as those resulting from the classical $K$-means algorithm. However, given that there is also a variable selection aspect to the clustering, these starting values may not perform well, especially when there are many uninformative variables. In order to incorporate the potential sparsity in the starting values, we use the following procedure:
\begin{enumerate}
\item Cluster the data using classical $K$-means, obtaining $K$ initial cluster centers $\bmu_{1, \cdot},\ldots, \bmu_{K, \cdot}$.
\item Compute the Euclidean norm for each variable center: $d_j = ||\bmu_{\cdot, j}||_2$ for $j = 1, \ldots, p$ and order them in descending order.
\item Execute $K$-means on the subset of variables corresponding to the $1, 2, 5, 10, 25$ and $50$ \% largest $d_j$.
\item Use the cluster indices of each of these $K$-means runs as an input for the regularized $K$-means version of Lloyd's algorithm, and choose the one yielding the lowest objective function.  
\end{enumerate}

\noindent The procedure outlined above allows the algorithm to start from several sparse solutions. The selection of the initial sparse solutions is based on the (Euclidean) norm of the variable centers, which is precisely what is penalized in regularized $K$-means clustering. This procedure is slightly slower, but does not increase the overall complexity of the algorithm.

\section{Comparison of penalty types}\label{sec:sim_pen}
We conduct a simulation study to compare the different penalty types and make a case for $\pen_0$, we perform a particular simulation study. In this simulation study, we give the algorithm two essential pieces of information. First, the true value of $K$ is kept fixed. Second, the solution to Equation \ref{eq:regkmeans} is computed on a grid of values for the regularization parameter $\lambda$, after which the partition that is closest to the theoretically correct partition is retained. While this is clearly a non-realistic setting as we need the true clustering in order to select the solution, the approach should give an idea of which penalty has the most potential, provided that the tuning parameters are chosen appropriately.\\

The data generation process starts from the approach of \cite{Sun2012} and extends this in several directions. We generate datasets of $n \in \{80, 800\}$ observations $\bx_1, \ldots, \bx_n$ in $p \in \{50, 200, 500, 1000\}$ dimensions. First the true cluster assignment vector $\by$ is sampled from $\{1, \ldots, K\}$ where $K \in \{2, 4, 8\}$. Then, for each observation $\bx_i$, the first 50 are the informative variables. They are sampled from $\mathcal{N}(\bmu(\by_i), I_{50})$, where $\bmu(\by_i)$ is given by 

\begin{align*}
\bmu_{K=2}(\by_i) &=
\mu\boldsymbol{1}_{50} I_{\by_i = 1}
-\mu\boldsymbol{1}_{50} I_{\by_i = 2}\\
\bmu_{K=4}(\by_i) &=
(-\mu \boldsymbol{1}_{25}, \mu \boldsymbol{1}_{25}) I_{\by_i = 1}+
\mu\boldsymbol{1}_{50} I_{\by_i = 2}+
(\mu \boldsymbol{1}_{25}, -\mu \boldsymbol{1}_{25}) I_{\by_i = 3}
-\mu\boldsymbol{1}_{50} I_{\by_i = 4}\\
\bmu_{K=8}(\by_i) &=
(\mu \boldsymbol{1}_{17}, \mu \boldsymbol{1}_{17}, \mu \boldsymbol{1}_{16} ) I_{\by_i = 1}+
(\mu \boldsymbol{1}_{17}, -\mu \boldsymbol{1}_{17}, \mu \boldsymbol{1}_{16} ) I_{\by_i = 2}\\
&+
(\mu \boldsymbol{1}_{17}, \mu \boldsymbol{1}_{17}, -\mu \boldsymbol{1}_{16} ) I_{\by_i = 3}+
(\mu \boldsymbol{1}_{17}, -\mu \boldsymbol{1}_{17}, -\mu \boldsymbol{1}_{16} ) I_{\by_i = 4}\\
&+
(-\mu \boldsymbol{1}_{17}, \mu \boldsymbol{1}_{17}, \mu \boldsymbol{1}_{16} ) I_{\by_i = 5}+
(-\mu \boldsymbol{1}_{17}, -\mu \boldsymbol{1}_{17}, \mu \boldsymbol{1}_{16} ) I_{\by_i = 6}\\
&+
(-\mu \boldsymbol{1}_{17}, \mu \boldsymbol{1}_{17}, -\mu \boldsymbol{1}_{16} ) I_{\by_i = 7}+
(-\mu \boldsymbol{1}_{17}, -\mu \boldsymbol{1}_{17}, -\mu \boldsymbol{1}_{16} ) I_{\by_i = 8}
\end{align*}

The parameter $\mu$ determines the separation of the clusters. When $\mu$ is large, the clusters are well separated, whereas a small value of $\mu$ will result in a lot of overlap between the clusters. We vary the value of $\mu$ in $\{0.4, 0.5, 0.6, 0.8\}$. To these informative variables, $p-50$ noise variables are added, which are sampled randomly from $\mathcal{N}(0,1)$. For each of the 24 simulation settings, we generate 100 datasets and average the results over these replications.\par
In order to evaluate clustering performance, we calculate the adjusted rand index (ARI) \citep{rand1971objective,hubert1985comparing} between the estimated partition and the true clustering. The ARI has an expected value of 0 for random clusterings of the data, whereas a perfect agreement corresponds with an ARI of 1.\par 
We compare the regularized $K$-means algorithms with the different penalties. Each of the methods is calculated on a grid of 40 lambda values given by $10^{-2 + 4i / 40}$, for $i = 0, 1, \ldots, 39$, after which the best solution is retained. We discuss the results for $n = 80$, $K = 4$ and $p=1000$ here. The results for the other settings are qualitatively similar and can be found in the Supplementary material. Table \ref{tab:comppen_p1000_start2} shows the results for this setting. We note that the scenarios with $\mu =0.4$ and $\mu = 0.5$ are very hard for all penalty types and none of them achieve a satisfactory performance. As the clusters get more separated, the penalized $K$-means starts to substantially outperform classical $K$-means. Out of the different penalty types, the ridge penalty is the least effective, whereas the hard-thresholding is most effective. The group-lasso is a close second, and the lasso penalty falls somewhere in between. 

\begin{table}[!ht]
\centering
\begin{tabular}{rllll}
  \hline
 & $\mu = 0.4$ & $\mu = 0.5$ & $\mu = 0.6$ & $\mu = 0.8$ \\ 
  \hline
classical & 0.08 (0.05) & 0.19 (0.08) & 0.35 (0.11) & 0.69 (0.12) \\ 
  ridge & 0.15 (0.05) & 0.28 (0.08) & 0.44 (0.11) & 0.75 (0.14) \\ 
  lasso & 0.15 (0.06) & 0.32 (0.1) & 0.69 (0.19) & 1 (0.02) \\ 
  glasso & 0.15 (0.06) & 0.36 (0.13) & 0.8 (0.19) & 1 (0) \\ 
  HT & 0.15 (0.06) & 0.34 (0.12) & 0.86 (0.17) & 1 (0) \\ 
   \hline
\end{tabular}
\caption{ARI values and standard deviations of regularized $K$-means variants on data of dimension $n = 80$ and $p =1000$ with $K = 4$ clusters.} 
\label{tab:comppen_p1000_start2}
\end{table}

\clearpage
\section{Consistency and variable selection}\label{sec:theo}

In order to investigate regularized $K$-means from a theoretical perspective, we consider the asymptotic  formulation of the objective function in Equation \ref{eq:regkmeans}. Let $Q$ be a probability measure on $\mathbb{R}^p$ and $A$ a finite subset of  $\mathbb{R}^p$. Let $\lambda \geq 0$ be fixed. We define the following objective function:
\begin{equation*}
W(A, Q) \coloneqq \int{\min_{a\in A}{||x-a||_2^2} Q(dx)}+  \lambda \pen(A)
\end{equation*}
where $\pen(A)$ denotes the penalization of the cluster centers in $A$.\par

Now fix $p$ and consider a $p$-variate random variable $X$ with distribution function $F$ and we fix an integer $K \geq 1$. 
\noindent Assume that
\begin{enumerate}[{a)}]
\item $\int{||x||_2^2 F(dx)} < \infty$
\item For each $k = 1, \ldots, K$, there is a unique set $\bar{A}(k)$ for which
\[W(\bar{A}(j), F) = \inf{\{W(A,F)|A \mbox{ contains at most $k$ points}\}}\]
\item $\pen$ is one of $\pen_0$, $\pen_1$, $\pen_2$ or $\pen_3$.
\end{enumerate}
These assumptions are identical to the assumptions needed for the consistency of classical $K$-means, see \cite{Pollard1981}. The following theorem establishes the (strong) consistency of regularized $K$-means in terms of the Hausdorff distance. For two finite sets $A$ and $B$, the Hausdorff distance between them is given by $$d_H(A,B) =\max\left\{\max_{a \in A}{\min_{b \in B}{||a-b||_2}}, \max_{b \in B}{\min_{a \in A}{||a-b||_2}}\right\}$$.

\begin{theorem}\label{theo:consistency}
Let $\bx_1 \ldots, \bx_n$ be a random sample from $F$ with empirical distribution function $F_n$ and let $A_n$ be optimal set of at most $K$ cluster centers for the sample. Under the conditions mentioned above, we have that:
\begin{enumerate}
\item $W(A_n, F_n) \xrightarrow{a.s.} W(\bar{A}(k), F)$
\item $A_n \xrightarrow{a.s.} \bar{A}(k)$
\end{enumerate}
where the convergence of sets is understood in terms of Hausdorff distance.
\end{theorem}

\noindent In addition to consistency, it would be nice to have some guarantee that the penalization works as intended, i.e. that we perform variable selection. Suppose w.l.o.g. that the last  $p-p_0+1$ variables are noise variables, in the sense that they are independent of all other variables and of any clustering structure. The optimal solution to the classical as well as the regularized $K$-means problem is then a set of centers $\bar{A}$ for which $\bar{A}_{\cdot, p_0:p} = 0$, i.e. the last $p-p_0+1$ centers are zero. We would then like to have $P(\hat{A}_{\cdot, j} = 0) \to 1$ for each $j= p_0, \ldots, p$. The strong consistency implies that the true zero-centers converge in probability to zero. However, this doesn't guarantee that the probability that they are equal to zero converges to 1, which is what we need to guarantee variable selection. The following theorem shows that this does indeed happen for all but the $\pen_2$ penalty, provided $\lambda >0$.

\begin{theorem}\label{theo:varsel}
Under the conditions above, and assuming that $\lambda > 0$, we have that $P(\hat{A}_{\cdot, j} = 0) \to 1$ for all  $j= p_0, \ldots, p$ for $\pen_0$, $\pen_1$ and $\pen_3$.
\end{theorem}

\noindent Ideally, one may wish that the non-zero centers are estimated as if the regular $K$-means algorithm would be executed on the ``selected'' variables, i.e. those variables with non-zero cluster centers. The following theorem shows that this can only happen for the $\pen_0$ penalty, provided the value of $\lambda$ is chosen correctly.

\begin{theorem}\label{theo:varsel2}
Under the conditions above, and using penalty $\pen_0$, there exists a $\lambda > 0$, such that we have that $P(\hat{A}_{\cdot, j} = 0) \to 1$ for all  $j= p_0, \ldots, p$ and $\hat{A}_{\cdot, j} \xrightarrow{P} A^*_{\cdot, j}$ for all  $j= 1, \ldots, p_0 - 1$. Where $A_*$ are the optimal cluster centers obtained by dropping the penalty term from the objective function (i.e. classical $K$-means).
\end{theorem}

We conjecture that Similar asymptotic results may be derived under more stringent conditions along the lines of the proof in \cite{Sun2012} for the case of diverging $p$.

\section{Selection of $\lambda$}\label{sec:lambdasel}
The selection of the regularization parameter $\lambda$ is not an easy task. The main reason is that many techniques and heuristics for tuning hyperparameters in cluster analysis rely on some kind of distance between observations. In the setting of regularization, one could calculate distances on the selected variables, or on all of the variables. In the former case, the distances are not comparable over different values of the regularization parameter. In the latter case, the values of the distances can be dominated by uninformative variables which are not used for the clustering. Therefore, relying on distances between observations may be inappropriate in the setting of regularized clustering. This makes straight forward adoption of popular methods such as the gap statistic \cite{tibshirani2001estimating}.
or the silhouette coefficient \citep{rousseeuw1987silhouettes} impossible.\\

We focus on the selection of $\lambda$ for the hard-thresholding penalty, which has the advantage of not shrinking the cluster centers of the selected variables. Through a simulation study, we will compare several options for selecting $\lambda$. We first consider the rather simple AIC and BIC \citep{ramsey2008uncovering} criteria given by

\begin{align*}
\mbox{AIC} &= \WCSS + 2 k p\\
\mbox{BIC} &= \WCSS + k \ln(n)p,
\end{align*}
where $\WCSS$ denotes the within-cluster sums of squares calculated on all variables. Possible improvements on these rather naive criteria could be in the form of more accurate estimation of the degrees of freedom in the BIC criteria, see \cite{HOFMEYR2020106974}.\\

In addition to the AIC and BIC criteria, we consider methods based on clustering stability rather than coherence-type measures. The idea in this approach is that a good clustering method should yield ``stable'' clusters, in the sense that it should yield similar cluster assignments when estimated on different samples from the same population. According to \cite{ben2006sober,Wang2010} we can define 
\begin{definition}[Clustering Distance]
The distance between any two clusterings $\psi_1$ and $\psi_2$ is defined as 
\begin{equation*}
d(\psi_1,\psi_2) = \Pr[I\{\psi_1(X) = \psi_1(Y)\} + I\{\psi_2(X) = \psi_2(Y)\}],
\end{equation*}
where $I(\cdot)$ denotes the indicator function and $X$ and $Y$ are independently sampled from $F$. 
\end{definition}

Based on the clustering distance above, we can define clustering instability as \citep{Wang2010}:
\begin{definition}[Clustering Instability]
The clustering instability of a clustering algorithm $\bm \psi$ is
\begin{equation*}
s(\bm \psi; \lambda) = \Exp[d\{\bm \psi(\bX_1; \lambda), \bm \psi(\bX_2; \lambda)\}],
\end{equation*}
where the expectation is taken with respect to $\bX_1$ and $\bX_2$ which are independent samples of size $n$ from $F$.  
\end{definition}

Several ways to estimate $s(\bm \psi; \lambda)$ have been proposed. \cite{Wang2010} propose to repeatedly split the data into 3 parts, 2 training sets and one validation set. The clustering method is trained on each of the training sets, and the stability is calculated as the expectation of their agreement in clustering the validation set. The problem with this approach is that the resulting datasets of sizes $n/3$ may be too small. We consider three alternatives:

\begin{enumerate}
\item \textbf{stab1:} \cite{Fang2012} propose instead to use bootstrap samples by taking for each replication, 2 bootstrap datasets of size $n$, after which the original data is used as validation set.
\item \textbf{stab2:} \cite{ben2001stability,haslbeck2020estimating} take the intersection of unique samples of the 2 bootstrapped training datasets as validation set.
\item \textbf{stab3:} \cite{Sun2012} use a variation where a third bootstrap dataset is taken as validation set. 
\end{enumerate}

In addition to the AIC, BIC and stability criteria, we also consider a different strategy for the selection of $\lambda$ which we call the \textit{gap method}. Suppose we have a certain clustering of the data based on $q$ variables with corresponding WCSS equal to $\WCSS_q$. Now, adding one variable to the dataset will lead to an increase of the WCCS equal to $\Delta_{q+1} \coloneqq \frac{1}{n}\WCSS_{q+1} - \frac{1}{n}\WCSS_q$. Because the variables are standardized, we have that $0\leq \Delta_{q+1} \leq 1$. Note that these bounds are sharp, as we have $\Delta_{q+1} = 0$ when the extra variable perfectly agrees with the clustering based on $q$ variables (i.e. the cluster assignments do not change) and additionally has a degenerate distribution (i.e. takes a fixed and different value for each cluster). $\Delta = 1$ is the increase to the WCSS when the cluster centers of the new variable are all put at zero. So, $\Delta_{q+1}$ characterizes how much the added variable agrees with the existing clustering based on $q$ variables. The closer it is to 1, the larger the disagreement. This makes it reasonable to look at the addition of variables to the model, and continue adding variables as long as the increase in WCSS is not too large. We do this by means of the following strategy which is somewhat similar in spirit to the Gap statistic of \cite{tibshirani2001estimating}.\\

Given the observed $\Delta_{q+1}$, the increase in WCSS for adding the $q+1$-th variable, we now need a basis for comparison in order to assess whether this increase is large. Assume without loss of generality that the variables enter the model in the order of their column number, i.e. first $q$ variables which entered the model correspond with the variables $x_{\cdot,1},\ldots, x_{\cdot,q}$. Now, we randomly permute the $q+1$-th variable to obtain $x_{\cdot, q+1}^*$, and we apply $K$-means to the dataset $x_{\cdot,1},\ldots, x_{\cdot,q}, x_{\cdot, q+1}^*$. The resulting WCSS is denoted $\WCSS_{q+1}^*$ and we also obtain a corresponding $\Delta_{q+1}^* \coloneqq \frac{1}{n}\WCSS_{q+1}^*-\frac{1}{n}\WCSS_{q}$. This procedure is repeated $S$ times, yielding estimates $m$ and $s^2$ for the expected value $E^*\left[\Delta_{q+1}^*\right]$ and variance $\var\left[\Delta_{q+1}^*\right]$ of the increase in WCSS caused by adding a random variable to the data. We take $S = 50$ in the rest of the paper. We now compare the observed with the expected delta (under randomness) yielding the value $D_{q+1} \coloneqq (m - \Delta_{q+1} )/s$. By computing these values along the sequence of active variables, we can select the lambda parameter yielding the set of active variables for which $D$ is maximal, or alternatively, the smallest lamba for which $D$ is within $c$ standard deviations of the maximum.\\

The strategy above works as long as the active set of variables are nested subsets. While this seems to be rather likely in smaller datasets, it is not guaranteed. Especially in larger datasets with many noise variables it happens that variables drop from the active set and re-enter at a later stage. In order to deal with this as well as with the entering of several variables at once, we slightly adapt the procedure. 
Suppose we have a grid of lambda values $\lambda_1 \geq \ldots \geq \lambda_L \geq 0$ with corresponding sets of active variables $\mathcal{A}_1,\ldots, \mathcal{A}_L$. We also have observed values of $\Delta(\mathcal{A}_j,\mathcal{A}_{j-1}) = \frac{1}{n}\WCSS_{\mathcal{A}_j} - \frac{1}{n}\WCSS_{\mathcal{A}_{j-1}}$. If $\mathcal{A}_{j-1} \subset \mathcal{A}_j$, we obtain reference values $\Delta^*(\mathcal{A}_j,\mathcal{A}_{j-1}) = \frac{1}{n}\WCSS_{\mathcal{A}_{j}}^*- \frac{1}{n}\WCSS_{\mathcal{A}_{j-1}}$ by applying $K$-means on the dataset consisting of the variables in $\mathcal{A}_{j-1}$ together with the randomly permuted variables in $\mathcal{A}_j \backslash \mathcal{A}_{j-1}$. In case there are variables which drop out of the active set, we compute $\Delta^*(\mathcal{A}_j \cup \mathcal{A}_{j-1},\mathcal{A}_{j-1})$ and $\Delta^*(\mathcal{A}_j \cup \mathcal{A}_{j-1},\mathcal{A}_j)$, the difference of which yield the reference values.  If $|\mathcal{A}_j \backslash \mathcal{A}_{j-1}| > 1$, we divide the reference values by $|\mathcal{A}_j \backslash \mathcal{A}_{j-1}|$.\\

\noindent We now compare the methods described above in a simulation study. We use the same simulation setup as before, with the difference that we now focus only on the hard-thresholding penalty. As before, we present the case of $K=4$, $n = 80$ and $p=1000$ here, and refer to the Supplementary Material for the other simulation results. For the stability based methods, 20 replications of the resampling strategy were used. Table \ref{tab:compsel_p1000_ARIs} presents the resulting ARI values. Note first of all that the simple AIC criterion is almost always among the best performing methods. Only for $\mu = 0.5$, the gap method seems to slightly outperform the AIC. While the BIC performs similar to the AIC in this setting, we found that the AIC is more consistent in situations with lower $p$, as can be seen from the additional simulation results in the Supplementary Material. The gap1 and gap2 methods perform similarly to the AIC criterion. The stability based selection techniques do not seem appropriate here, and only start performing reasonably in the simplest case of $\mu = 0.8$.

\begin{table}[ht]
\centering
\begin{tabular}{rllll}
  \hline
 & $\mu = 0.4$ & $\mu = 0.5$ & $\mu = 0.6$ & $\mu = 0.8$ \\ 
  \hline
AIC & 0.09 (0.06) & 0.26 (0.12) & 0.8 (0.19) & 1 (0.01) \\ 
  BIC & 0.05 (0.05) & 0.21 (0.12) & 0.79 (0.24) & 1 (0) \\ 
  gap1 & 0.09 (0.06) & 0.28 (0.12) & 0.76 (0.23) & 0.99 (0.03) \\ 
  gap2 & 0.09 (0.05) & 0.27 (0.12) & 0.77 (0.23) & 0.99 (0.03) \\ 
  stab1 & 0.08 (0.05) & 0.22 (0.08) & 0.47 (0.21) & 0.82 (0.17) \\ 
  stab2 & 0.08 (0.05) & 0.2 (0.09) & 0.47 (0.23) & 0.84 (0.17) \\ 
  stab3 & 0.08 (0.06) & 0.21 (0.09) & 0.48 (0.21) & 0.84 (0.2) \\ 
   \hline
\end{tabular}
\caption{ARIs of lambda selection techniques
                                on data of dimension p = 1000} 
\label{tab:compsel_p1000_ARIs}
\end{table}

\begin{table}[ht]
\centering
\begin{tabular}{rllll}
  \hline
 & $\mu = 0.4$ & $\mu = 0.5$ & $\mu = 0.6$ & $\mu = 0.8$ \\ 
 \hline
AIC & 99.63 (18.12) & 98.87 (25.89) & 81.39 (23.63) & 90.42 (7.02) \\ 
  BIC & 6.06 (2.61) & 11.47 (6.2) & 35.76 (10.26) & 50.01 (0.95) \\ 
  gap1 & 99.51 (47.69) & 68.6 (49.15) & 34.51 (12.73) & 30.38 (7.4) \\ 
  gap2 & 121.48 (58.67) & 83.68 (60.74) & 35.47 (13.15) & 30.56 (7.56) \\ 
  stab1 & 486.64 (316.6) & 502.23 (303.41) & 490.77 (298.99) & 541.95 (285.42) \\ 
  stab2 & 502.69 (317.82) & 517.64 (313.86) & 462.54 (314.2) & 503.15 (301.03) \\ 
  stab3 & 495.89 (333.37) & 522.97 (313.2) & 483.47 (309.55) & 466.88 (301.17) \\ 
   \hline
\end{tabular}
\caption{Number of selected variables of lambda selection techniques
                                on data of dimension p = 1000} 
\label{tab:compsel_p1000_nbvars}
\end{table}

In addition to the resulting ARI values, we consider the number of selected variables. Remember that the true number of informative variables is 50. Table \ref{tab:compsel_p1000_nbvars} shows the number of selected variables for the scenarios under consideration. It is clear that it is quite difficult to select the correct number of variables. BIC seems to be closest to the true number of informative variables, but only when the cluster centers are very well separated. AIC appears to consistently overestimate the true number of informative variables, but as discussed before, this does not seem to lead to inferior performance in terms of recovering the class memberships. The gap methods select sparse solutions when the clusters are fairly well separated, whereas they tend to overestimate the number of informative variables otherwise. The stability-based methods select way too many variables, again suggesting that they are not preferable to use in combination with the hard-thresholding penalty.\\

Finally, we briefly consider the computation times of the different methods. It is clear that AIC and BIC are very quick to compute, since they require virtually no additional calculations once the regularized $K$-means algorithm has been executed on a grid of values for $\lambda$. The gap methods require substantially more computation time, and the stability-based criteria are very slow. This is of course due to the repeated runs of regularized $K$-means all the bootstrap samples.

\begin{table}[ht]
\centering
\begin{tabular}{rllll}
  \hline
 & $\mu = 0.4$ & $\mu = 0.5$ & $\mu = 0.6$ & $\mu = 0.8$ \\ 
  \hline
AIC & 0.23 (0.01) & 0.23 (0.02) & 0.23 (0.02) & 0.23 (0) \\ 
  BIC & 0.17 (0.01) & 0.17 (0.01) & 0.17 (0) & 0.18 (0) \\ 
  gap1 & 408.85 (32.09) & 382.48 (49.28) & 317.39 (60.25) & 169.02 (32.91) \\ 
  gap2 & 408.85 (32.09) & 382.48 (49.28) & 317.39 (60.25) & 169.02 (32.91) \\ 
  stab1 & 877.74 (32.35) & 886.47 (28.95) & 917.88 (30.8) & 943.94 (32.51) \\ 
  stab2 & 877.74 (32.35) & 886.47 (28.95) & 917.88 (30.8) & 943.94 (32.51) \\ 
  stab3 & 877.74 (32.35) & 886.47 (28.95) & 917.88 (30.8) & 943.94 (32.51) \\ 
   \hline
\end{tabular}
\caption{Computation time of lambda selection techniques
                                on data of dimension p = 1000} 
\label{tab:compsel_p1000_time}
\end{table}

We end the discussion of the simulation study with a remark. While automatic selection of the regularization parameter is attractive, we have found that in practice, there seems not to be any one-size-fits all method. Therefore, we encourage applying several methods and comparing the conclusions and results. We will illustrate this process in Section \ref{sec:examples} with real data examples.

\clearpage
\section{Simulation study}\label{sec:sim_comp}
We now compare HT $K$-means with the most popular competitors. The best-known competitor is the \textit{sparse $K$-means} method of \cite{Witten2010}. Sparse $K$-means was shown to outperform several alternative approaches such as the COSA method \citep{friedman2004clustering}, the model-based clustering of \cite{raftery2006variable} and PCA followed by $K$-means. We use the implementation of sparse $K$-means provided in the \texttt{R}-package \texttt{sparcl} by \cite{Witten2018}. The tuning parameter is chosen by the proposed permutation approach using 20 permutations searching over a grid of length 40. We additionally include the regularized $K$-means (henceforth Reg $K$-means) of \cite{Sun2012}, where the tuning parameter is chosen using the proposed stability criterion over 20 bootstrap replications and a grid of 40 lambda values given by $10^{-2 + 4i / 40}$, for $i = 0, 1, \ldots, 39$. Finally, we compare with classical $K$-means which serves as a reference.\\

Table \ref{tab:compcomp_p1000_ARIs} presents the ARI results on data of dimension $p = 1000$, where several things can be noted. First, we see that the cases of little separation between the cluster centers ($\mu = 0.4$ and $\mu = 0.5$) are really difficult, and none of the methods has a satisfactory performance. As the clusters get more separated, the clustering task clearly becomes easier. The peformance of HT $K$-means is never worse than that of the competitors, and substantially better in the case of $\mu = 0.6$ and $\mu = 0.8$. Sparse $K$-means is the second best performing method, with very competitive performance for $\mu = 0.8$ and a reasonable performance for $\mu = 0.6$. Reg $K$-means does not seem to be doing much better than classical $K$-means in this simulation. An important element in the explanation for this behavior is the fact that Reg $K$-means uses classical $K$-means as a starting value. Therefore, the method can more easily get suck in a local minimum which is close to its starting value, the classical $K$-means solution. Finally, note that classical $K$-means starts to perform reasonably well as the cluster centers get more and more separated.

\begin{table}[ht]
\centering
\begin{tabular}{rllll}
  \hline
 & $\mu = 0.4$ & $\mu = 0.5$ & $\mu = 0.6$ & $\mu = 0.8$ \\ 
  \hline
HT $K$-means & \textbf{0.09} (0.06) & \textbf{0.26} (0.12) & \textbf{0.8} (0.19) & \textbf{1} (0.01) \\ 
  Reg $K$-means & \textbf{0.09} (0.05) & 0.19 (0.08) & 0.36 (0.12) & 0.72 (0.14) \\ 
  Sparse $K$-means & 0.05 (0.05) & 0.18 (0.1) & 0.66 (0.27) & 0.96 (0.08) \\ 
  $K$-means & \textbf{0.09} (0.05) & 0.19 (0.08) & 0.36 (0.11) & 0.69 (0.12) \\ 
   \hline
\end{tabular}
\caption{ARIs of competitor comparison
                                on data of dimension p = 1000} 
\label{tab:compcomp_p1000_ARIs}
\end{table}

We now briefly consider the number of selected variables for each of the methods, shown in Table \ref{tab:compcomp_p1000_nbvars}. The AIC criterion used for HT $K$-means consistently underestimates the sparsity of the signal and selects a few too many variables on average. However, out of all the methods, it is closest to the true number of signal variables (50) most of the time. Sparse $K$-means seems to select too many variables when the clusters are not very well separated. However, for well-separated clusters ($\mu = 0.8$), it selects almost exactly 50 variables. Finally, Reg $K$-means heavily underestimates the sparsity of the signal. Again, the most likely cause is the fact that the algorithm starts from the classical $K$-means solution, which evidently uses all variables for clustering.

\begin{table}[ht]
\centering
\begin{tabular}{rllll}
  \hline
 & $\mu = 0.4$ & $\mu = 0.5$ & $\mu = 0.6$ & $\mu = 0.8$ \\ 
  \hline
HT $K$-means & 99.63 (18.12) & 98.87 (25.89) & 81.39 (23.63) & 90.42 (7.02) \\ 
  Reg $K$-means & 707.79 (242.45) & 724.07 (239.49) & 719.3 (252.14) & 654.38 (274.07) \\ 
  Sparse $K$-means & 236.83 (330.52) & 293.07 (305.69) & 130.51 (208.21) & 49.65 (92.76) \\ 
  $K$-means & 1000 (0) & 1000 (0) & 1000 (0) & 1000 (0) \\ 
   \hline
\end{tabular}
\caption{Number of selected variables competitor comparison
                                on data of dimension p = 1000} 
\label{tab:compcomp_p1000_nbvars}
\end{table}

Finally we take a brief look at the computation times of the different methods. Table \ref{tab:compcomp_p1000_time} shows the computation times in seconds. It is immediately clear that classical $K$-means is by far the fastest method and the regularized alternatives have a computation time that is larger by several orders of magnitude. Of these alternatives, HT $K$-means is the fastest to compute, followed by Sparse $K$-means which is about 4 times as slow on this data. Reg $K$-means is much slower than the competitors, and the bulk of this computation time is due to the stability-based tuning of the regularization parameter $\lambda$.

\begin{table}[ht]
\centering
\begin{tabular}{rllll}
  \hline
 & $\mu = 0.4$ & $\mu = 0.5$ & $\mu = 0.6$ & $\mu = 0.8$ \\ 
  \hline
HT $K$-means & 28.1 (1.01) & 28.22 (0.97) & 28.71 (0.95) & 27.96 (0.75) \\ 
  Reg $K$-means & 1990.84 (61.04) & 1999.47 (54.59) & 1983.21 (61.33) & 1999.58 (59.61) \\ 
  Sparse $K$-means & 100.31 (2.06) & 100.29 (2.52) & 100.7 (2.84) & 100.75 (2.23) \\ 
  $K$-means & 0.63 (0.02) & 0.63 (0.02) & 0.63 (0.02) & 0.6 (0.02) \\ 
   \hline
\end{tabular}
\caption{Computation time of competitor comparison
                                on data of dimension p = 1000} 
\label{tab:compcomp_p1000_time}
\end{table}


\clearpage
\section{Real data examples}\label{sec:examples}
In this section we analyze several real data examples using the HT $K$-means method. We start with a few simple examples and turn to more complex datasets after.

\subsection{The Iris dataset}
\begin{figure}[!h]
\includegraphics[width = 0.5\textwidth]{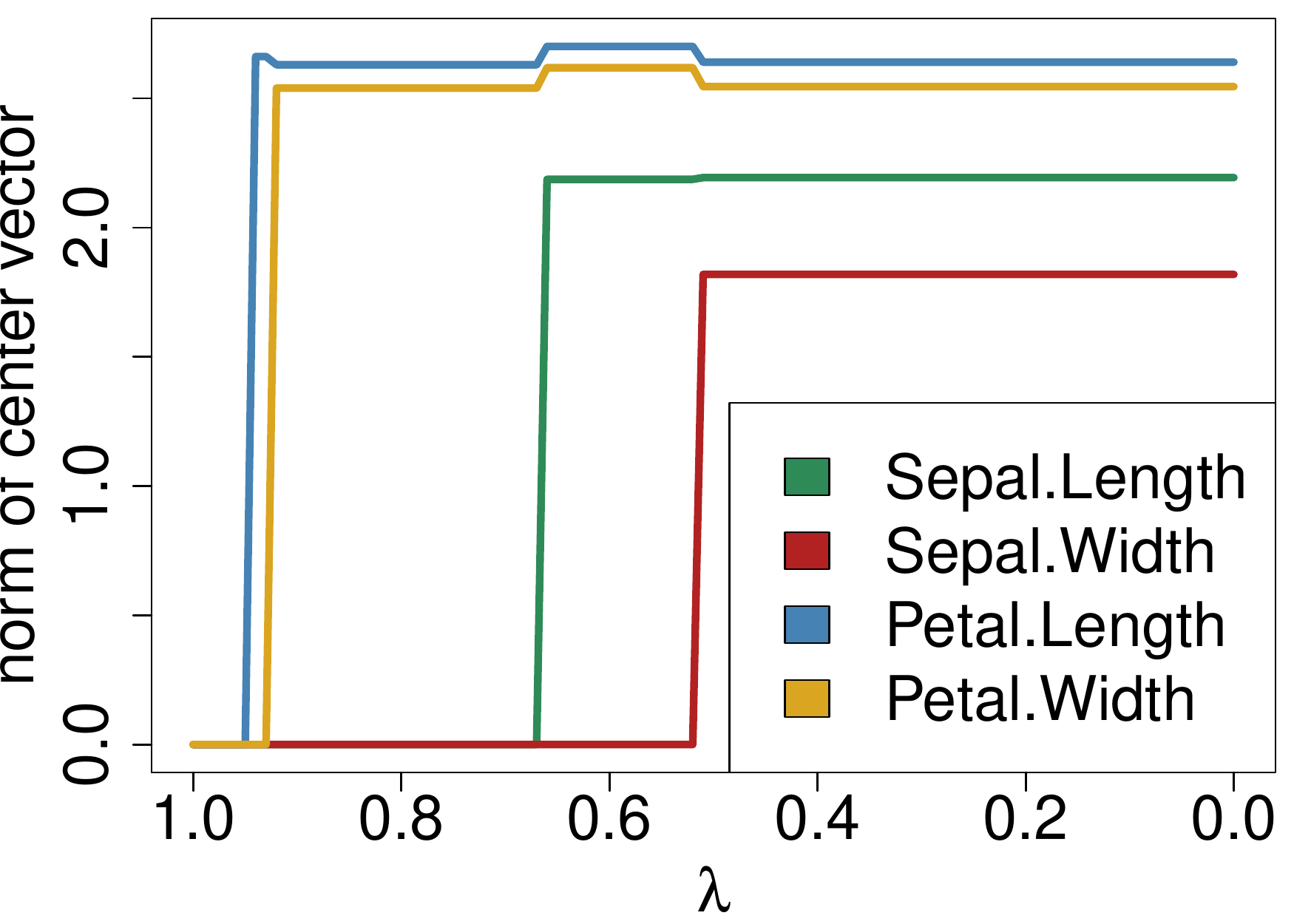}
\caption{Regularization path of HT $K$-means on the Iris data.}
\label{fig:iris_path}
\end{figure}

We first reconsider the Iris dataset discussed in the introduction, where it was clear that not all variables contribute equally to the partitioning of the data. More specifically, the third and fourth variable contain most information with respect to the true clustering structure. Adding more information does not help in recovering the underlying clustering, and in fact worsens the result. When applying HT $K$-means to the data, we obtain the regularization path of Figure \ref{fig:iris_path}. As $\lambda$ decreases, we see the the variables enter the active set of clustering variables one by one. As discussed in the introduction, the best clustering performance is achieved when using only the dimensions of the petal as information, i.e. the yellow and blue variables in the regularization path. Including the red and/or green variable worsens the result. 

The AIC and BIC criteria select a $\lambda$ parameter of 0, suggesting that all four variables should be used to cluster the Iris data. The stability based methods as well as the gap method select $\lambda$ between 0.67 and 0.92, meaning that they all select the 2 variables describing the dimensions of the petal and thus achieve the optimal ARI on this dataset.

\subsection{The banknote dataset}
\begin{figure}[!h]
\includegraphics[width = 0.65\textwidth]{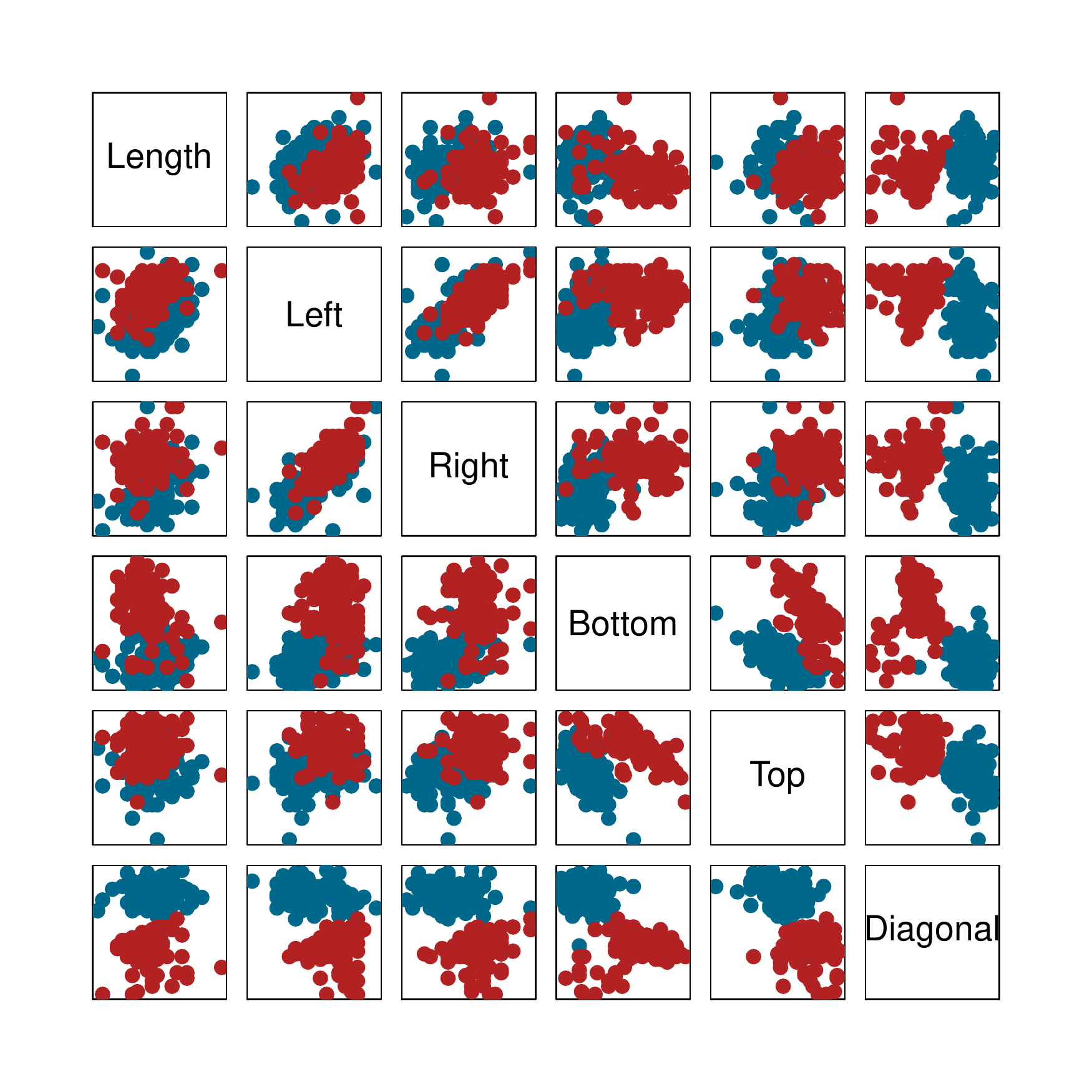}
\caption{Pairs plot of the banknote data. The genuine bank notes are colored in blue, whereas the counterfeit notes correspond with the red dots.}
\label{fig:banknote_pairs}
\end{figure}

As an additional small example we consider the banknote dataset, which consists of six measurements for 100 genuine and 100 counterfeit old-Swiss 1000-franc bank notes. The data was analyzed in \cite{flury1988multivariate} and is publicly available in the \texttt{R}-package \texttt{mclust} \citep{scrucca2016}.
For each bank note, we have the length, the width of the left and right edges, the bottom and top margin widths and the length of the diagonal. Figure \ref{fig:banknote_pairs} presents a pairs plot of the data, with the genuine and counterfeit bills colored in blue and red respectively. From this plot we may expect that not all variables contribute equally to the separation between good and bad bank notes. 

\begin{figure}[!h]
\includegraphics[width = 0.5\textwidth]{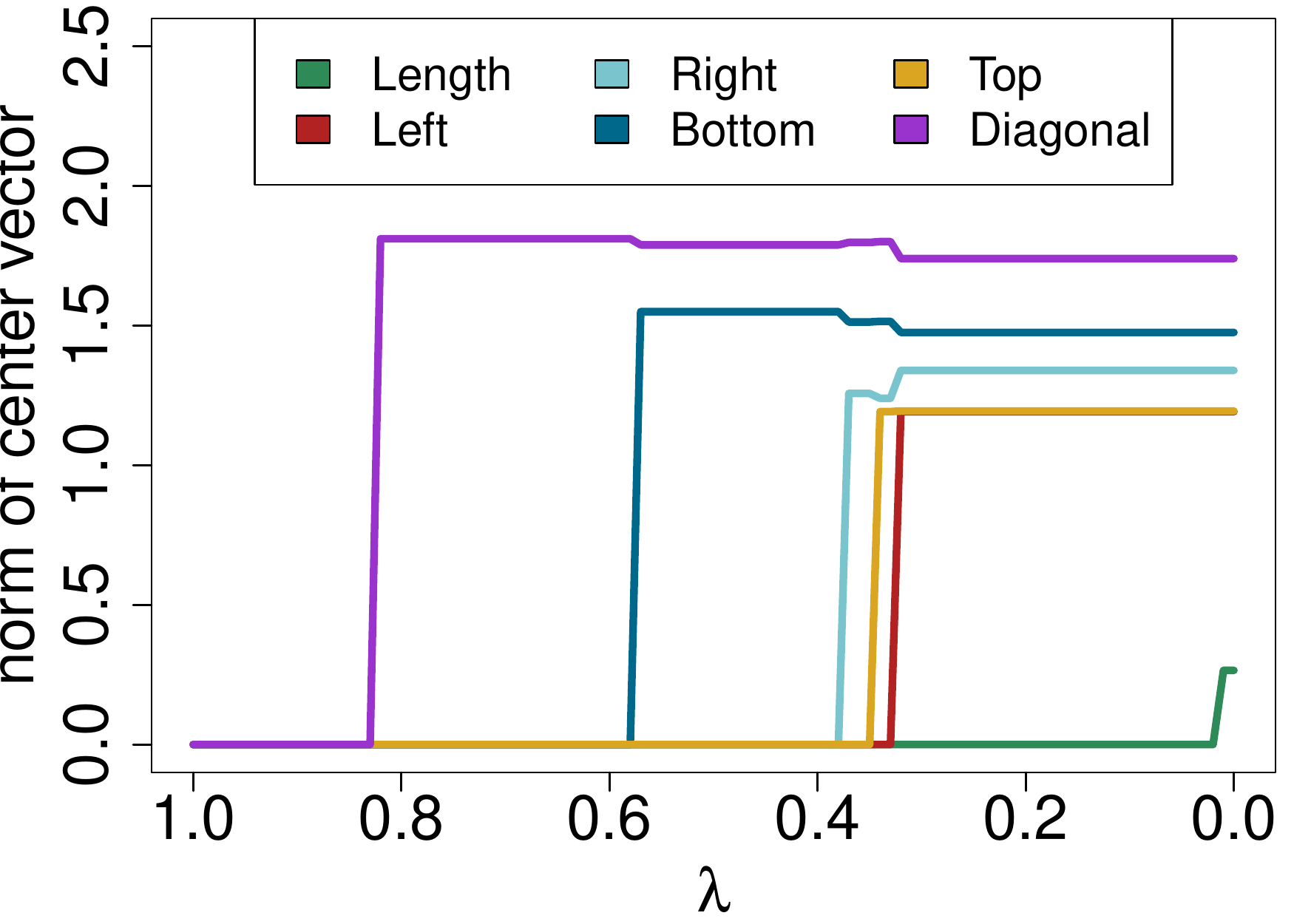}
\caption{Regularization path of HT $K$-means on the bank note data.}
\label{fig:banknote_path}
\end{figure}

If we cluster the bank note data using classical $K$-means, we obtain an ARI of about 0.85, which is already quite a good performance. Figure \ref{fig:banknote_path} shows the regularization path resulting from applying HT $K$-means on the bank note data. From this plot, we immediately see that not all variables contribute equally to the clustering of the data. More specifically, it seems that the measurements of the diagonal of the bill is by far the most important variable, followed by the bottom margin variable. It turns out that if we cluster only based on the diagonal measurement, we obtain an ARI of 96 \%. If we additionally include the second variable, the bottom margin, we obtain an ARI of 98 \%, which is almost perfect recovery of the true clusters. Including additional variables slightly lowers the ARI, but it is the measurements of the length and left edge which make the ARI drop from around 0.95 to 0.85. Any $\lambda$ value smaller than 0.33 includes these variables and thus we would like to select a tuning parameter value of at least 0.33. The AIC and BIC criteria select a $\lambda$ of 0.02, meaning that they leave out the length variable, but still include the left edge variable. The stability based criteria also select $\lambda$ values between 0.02 and 0.33, essentially selecting 5 variables and yielding a suboptimal ARI. The gap method selects a $\lambda$ value of 0.38, including only 2 variables in the clustering and achieving the optimal ARI on this dataset.

\subsection{The colon cancer dataset}
We analyze the gene expression data which is publicly available in the \texttt{R}-package \texttt{antiProfilesData} \citep{antiProfilesData2020} and contains samples of normal colon tissue and colon cancer tissue collected from the Gene Expression Omnibus \citep{GEO2002, GEO2012}. The complete dataset contains 68 gene expressions of length 5339, subdivided into 4 categories: adenoma, colorectal cancer, normal and tumor. There are 15 observations for each of the first three categories, and 23 of the tumor category. We are interested in clustering the data and thereby hopefully recovering (some of) the different tissue types in the obtained partition. Furthermore, if we are able to do so using a limited number of variables, this would benefit the insight gained into potentially important features.\par

\begin{figure}[!h]
\includegraphics[width = 0.75\textwidth]{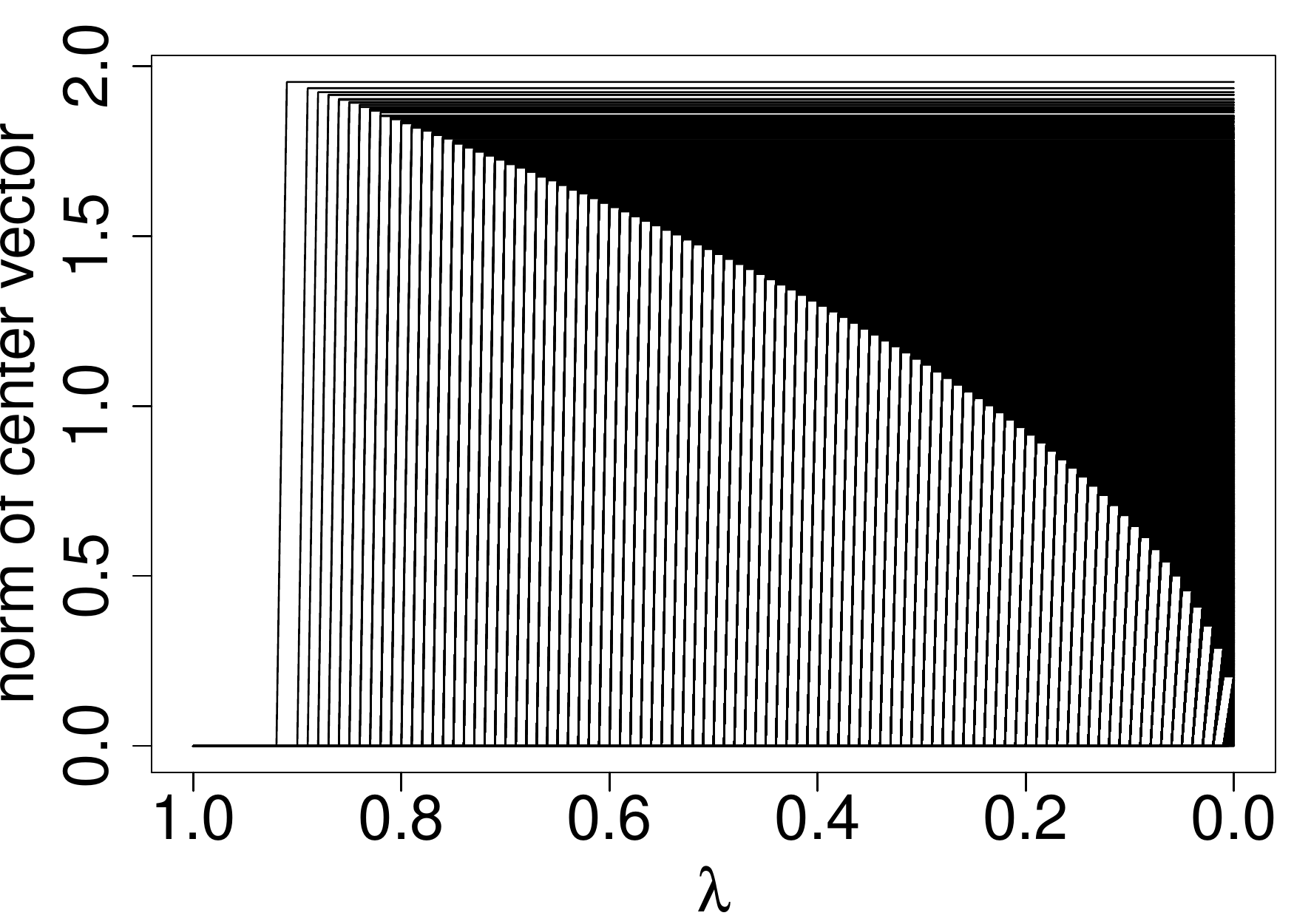}
\caption{Regularization path of HT $K$-means on the colon cancer data.}
\label{fig:coloncancer_path}
\end{figure}

We first consider the simplified problem of separating the normal tissue from the tumor, which together form a dataset of size $38 \times 5339$. Interestingly, when clustering this data using classical $K$-means, we obtain a perfect recovery of the true clusters: normal tissue vs. tumor tissue. However, classical $K$ means evidently uses all variables to obtain this partition, and offers no insight into whether all of these variables are needed or whether some of them may be redundant. Figure \ref{fig:coloncancer_path} shows the regularization path resulting from applying HT $K$-means to the colon cancer data. Clearly, not all variables contribute equally to the clustering, as even for very small values of the regularization parameter $\lambda$, many variables are dropped from the clustering. As even classical $K$-means clusters this data perfectly, we cannot hope to perform better in that respect, but we can try to identify potentially interesting features as well as try to obtain the same perfect partition using fewer variables. 

\begin{figure}[!h]
\includegraphics[width = 0.49\textwidth]{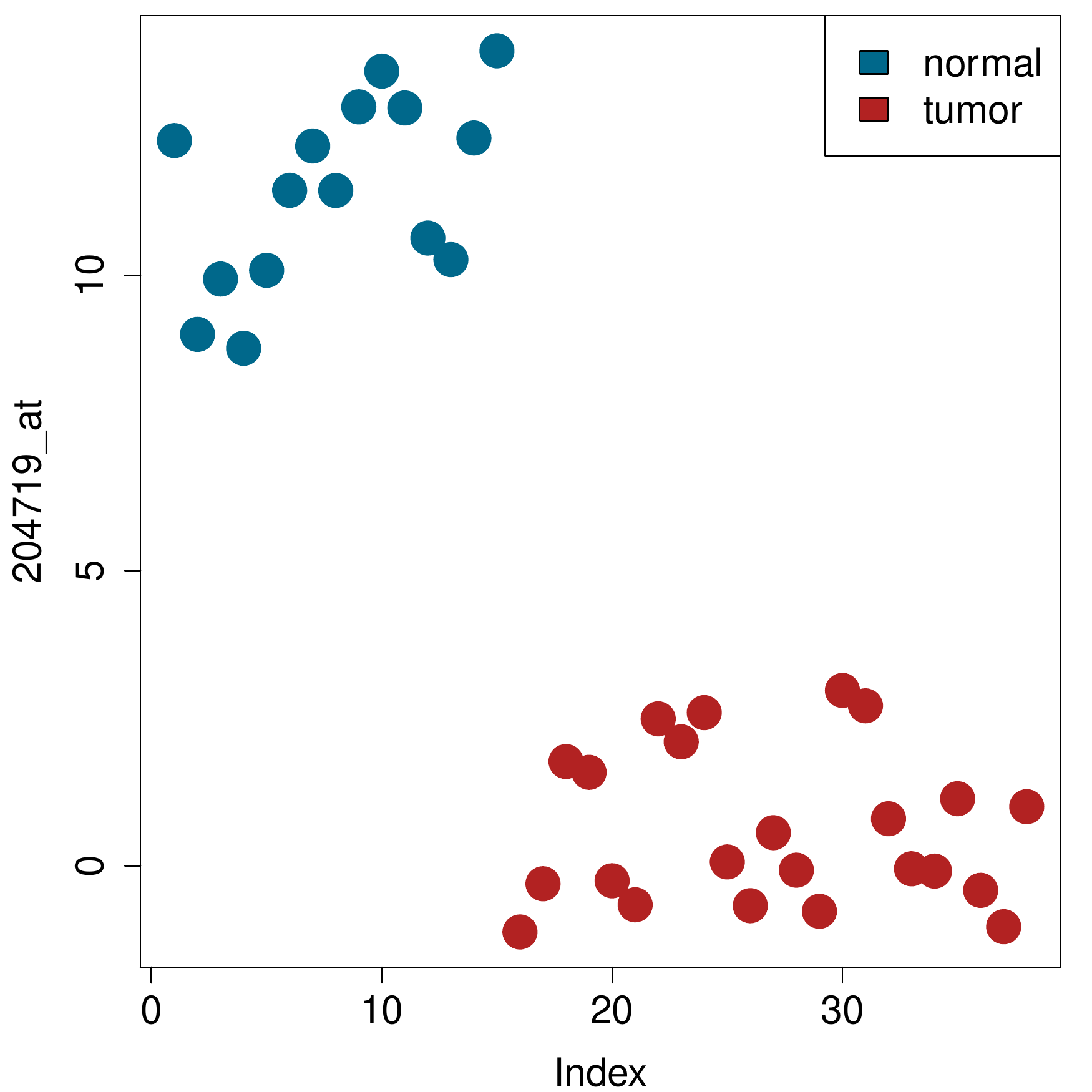}
\includegraphics[width = 0.49\textwidth]{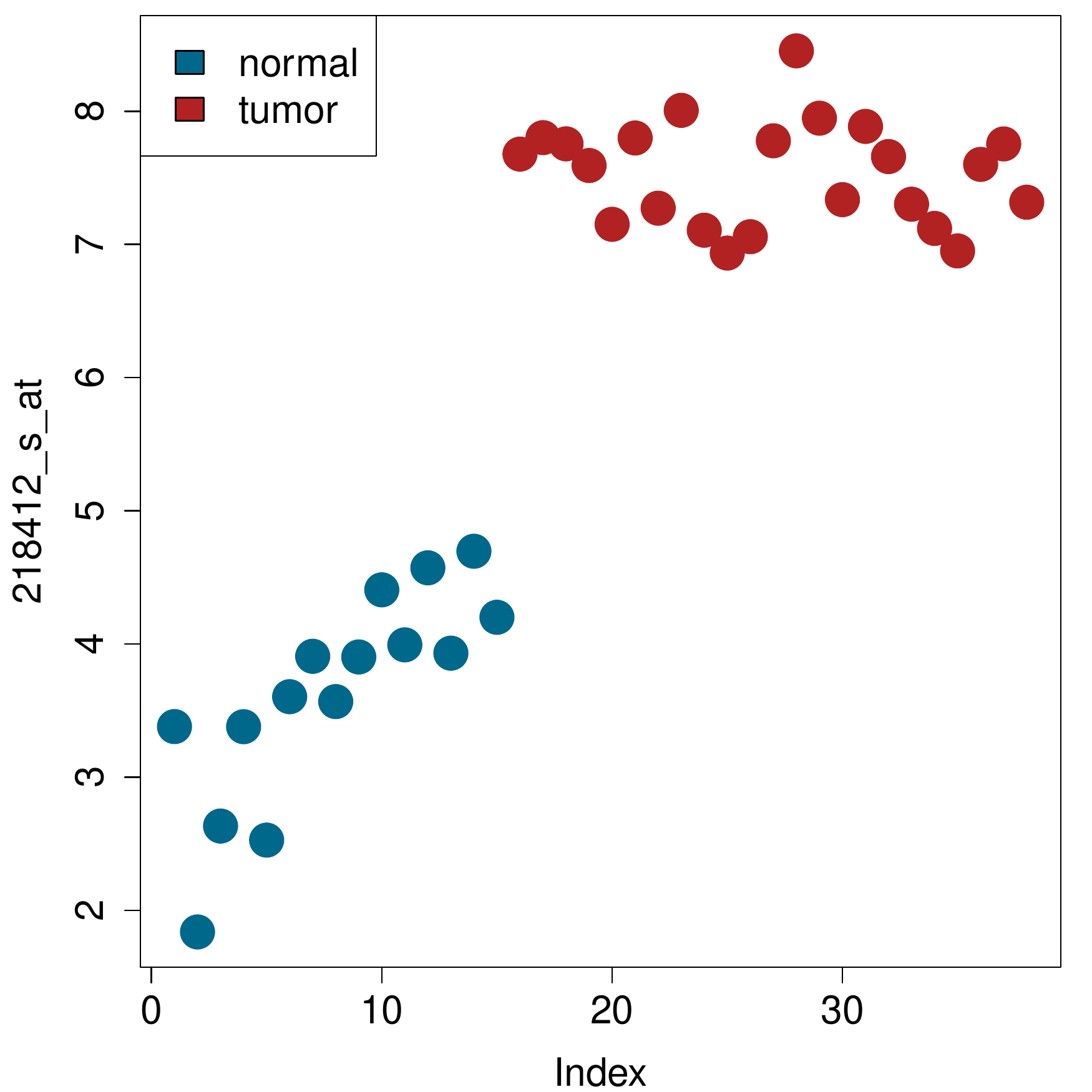}
\includegraphics[width = 0.49\textwidth]{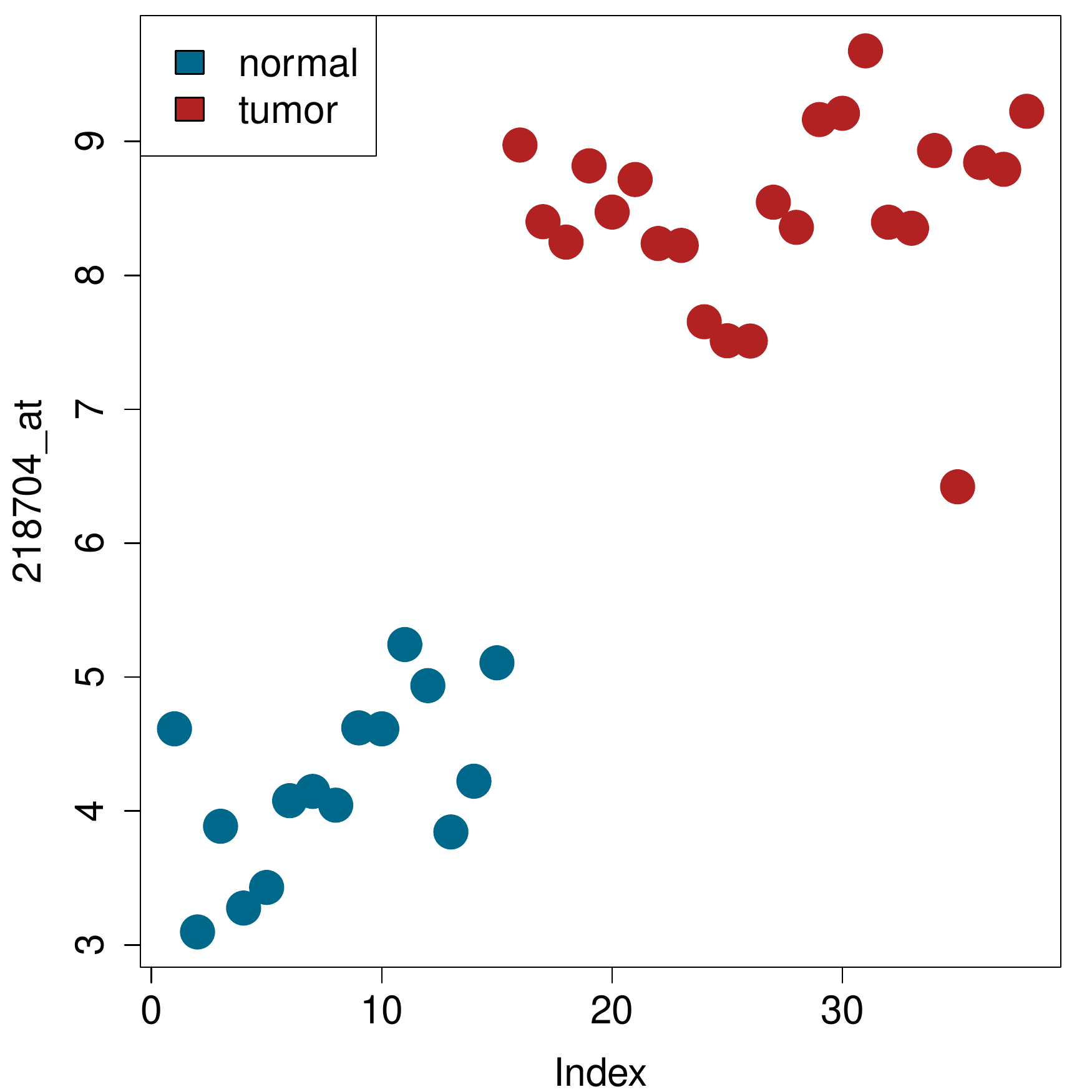}
\includegraphics[width = 0.49\textwidth]{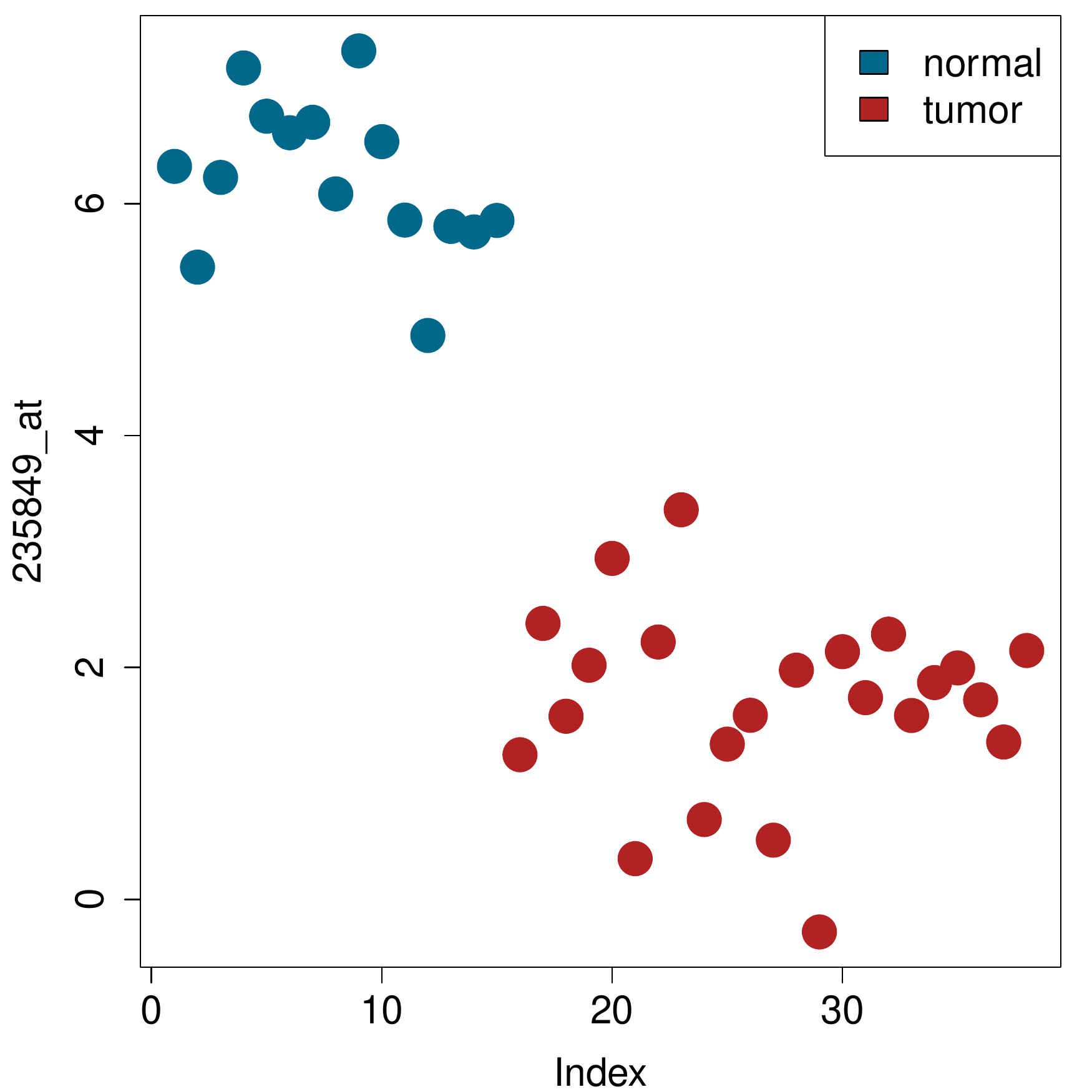}
\caption{Expression levels of the first 4 features entering the clustering model. All four variables perfectly separate the normal from the tumor samples.}
\label{fig:coloncancer_firstvariables}
\end{figure}

AIC and BIC suggest tuning parameters of 0.11 and 0.19 respectively, which yield clustering models of size 2697 and 1999. The gap method suggests using $\lambda = 0.62$, which corresponds to clustering based on just 277 variables. The stability based methods stab1, stab2 and stab3 suggest using $\lambda$ equal to 0.14, 0.08 and 0.26 respectively which correspond with clustering models of size  2396, 3046 and 1530 respectively. All of these options achieve perfect clustering, but the gap method selects the most sparse model. In fact, all sub models along the regularization path achieve perfect clustering, which makes it interesting to consider the first few variables which enter the active set of clustering features. Figure \ref{fig:coloncancer_firstvariables} shows the expression levels of the first 4 variables which enter the clustering model. All 4 of these variables perfectly separate the 
normal tissue samples from the tumor samples, explaining why the data is rather easy to cluster regardless of the tuning parameter. However, HT $K$-means allows us to identify these variables as they appear first in the regularization path.

\begin{figure}[!h]
\includegraphics[width = 0.49\textwidth]{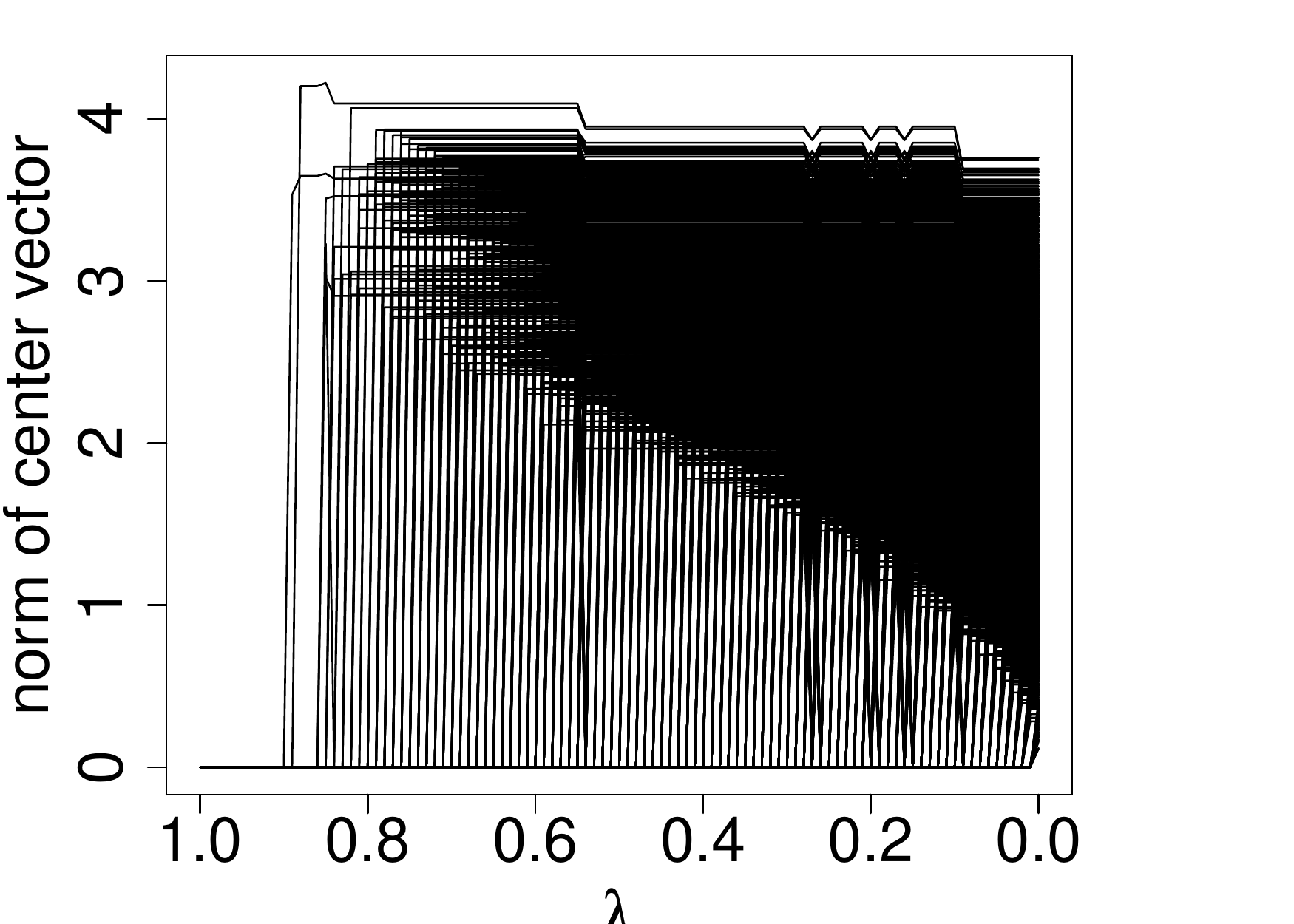}
\includegraphics[width = 0.49\textwidth]{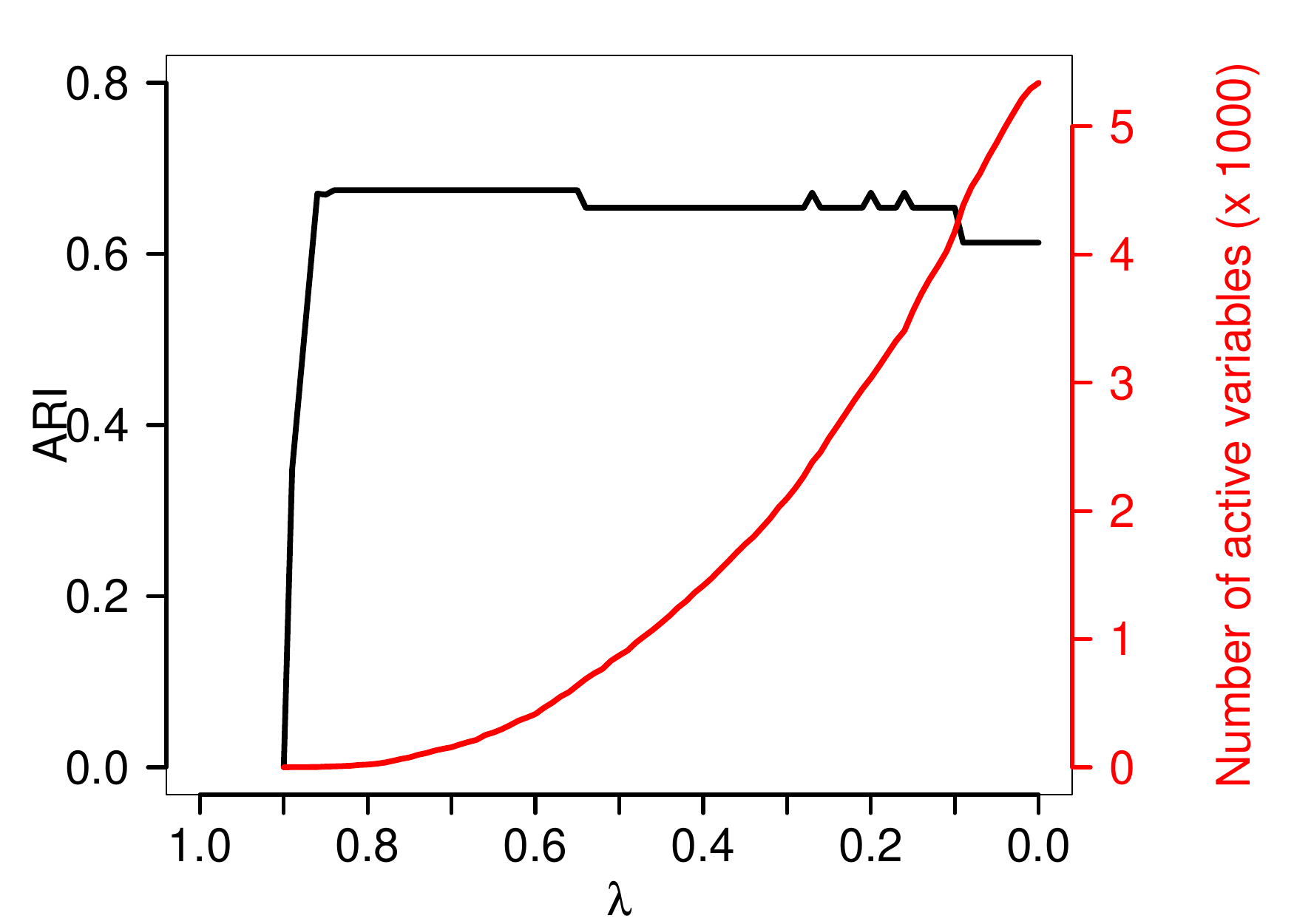}
\caption{Regularization path of the full colon cancer dataset (left) and a plot with the model size and ARI (right).}
\label{fig:coloncancerFull_path}
\end{figure}

\begin{figure}[!h]
\includegraphics[width = 0.49\textwidth]{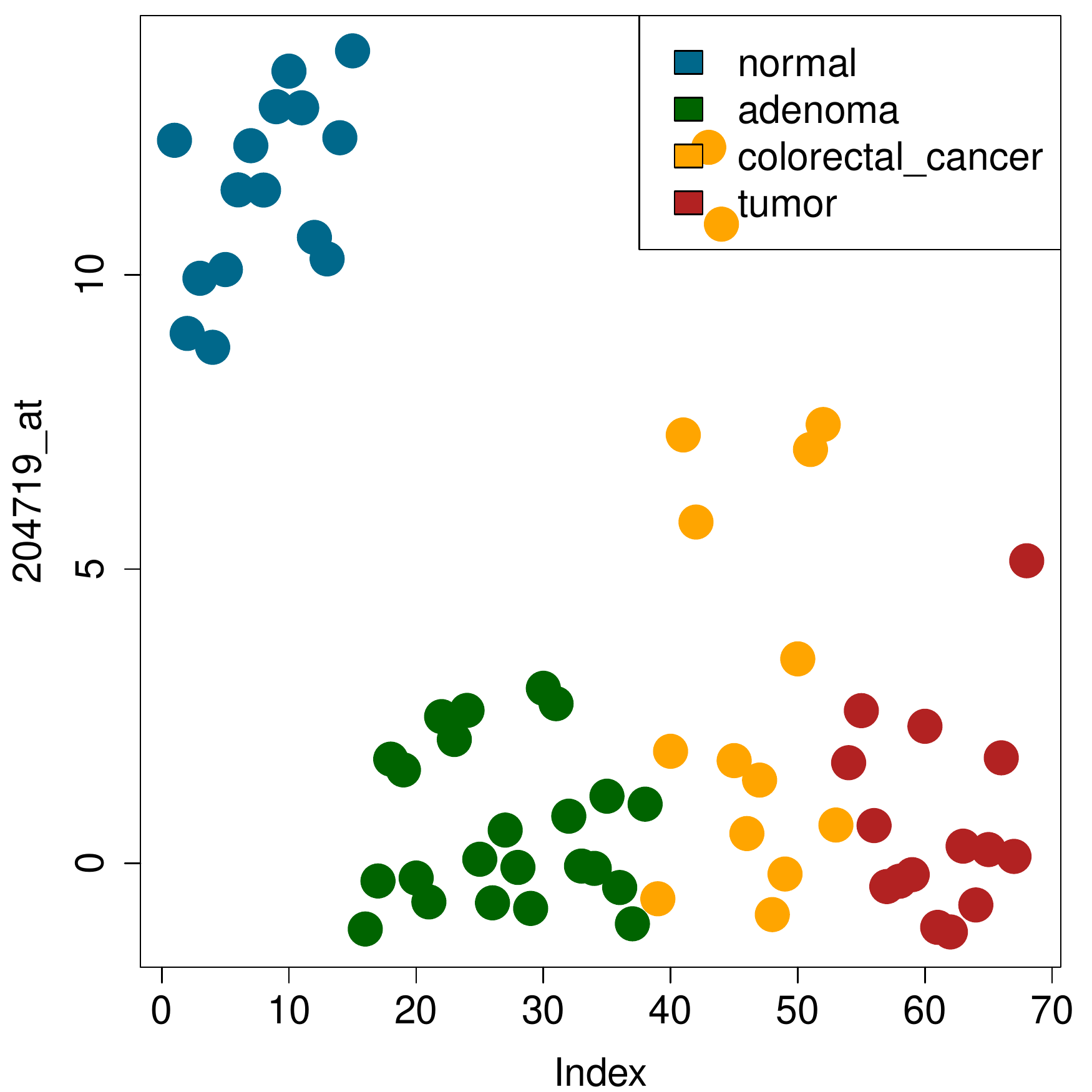}
\includegraphics[width = 0.49\textwidth]{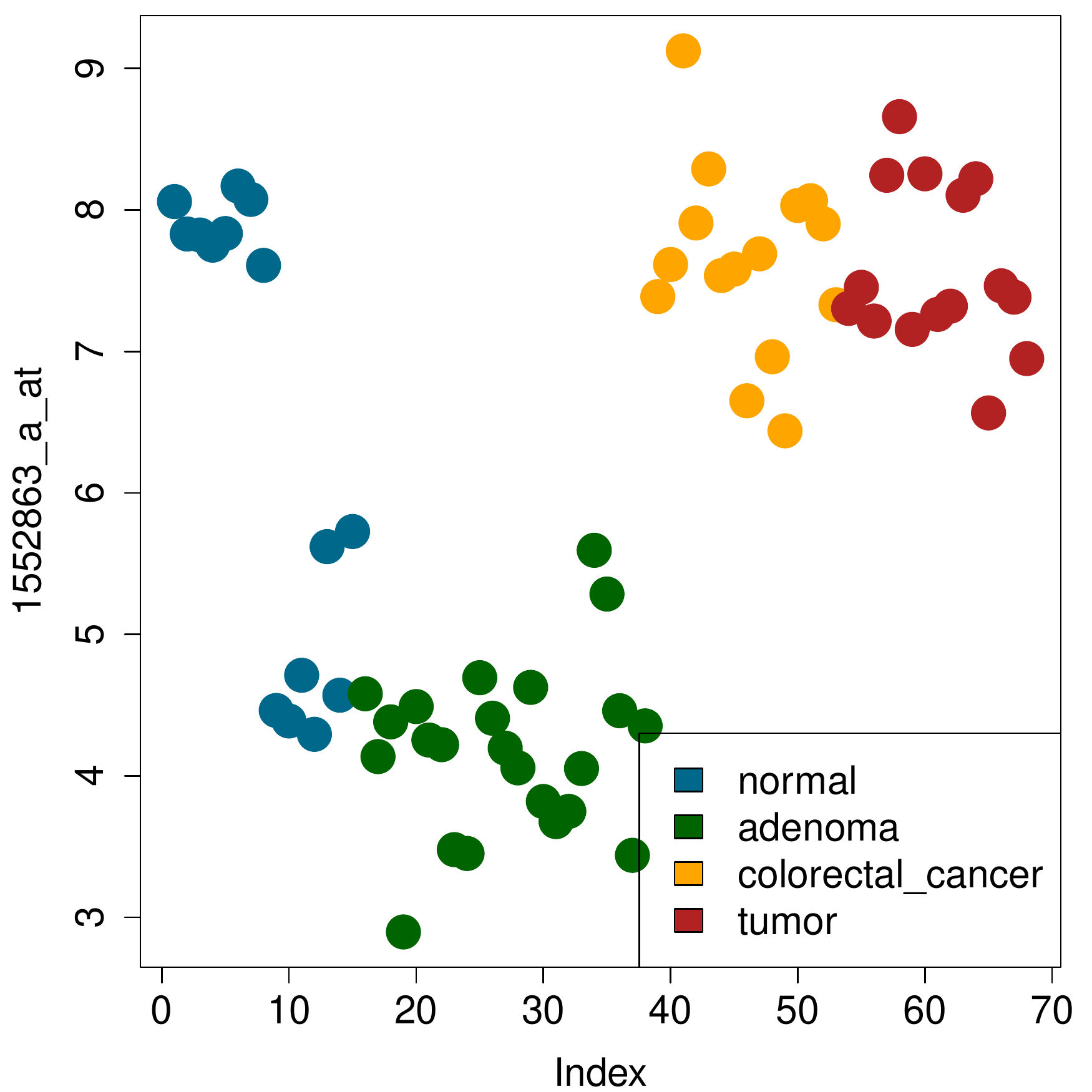}
\includegraphics[width = 0.49\textwidth]{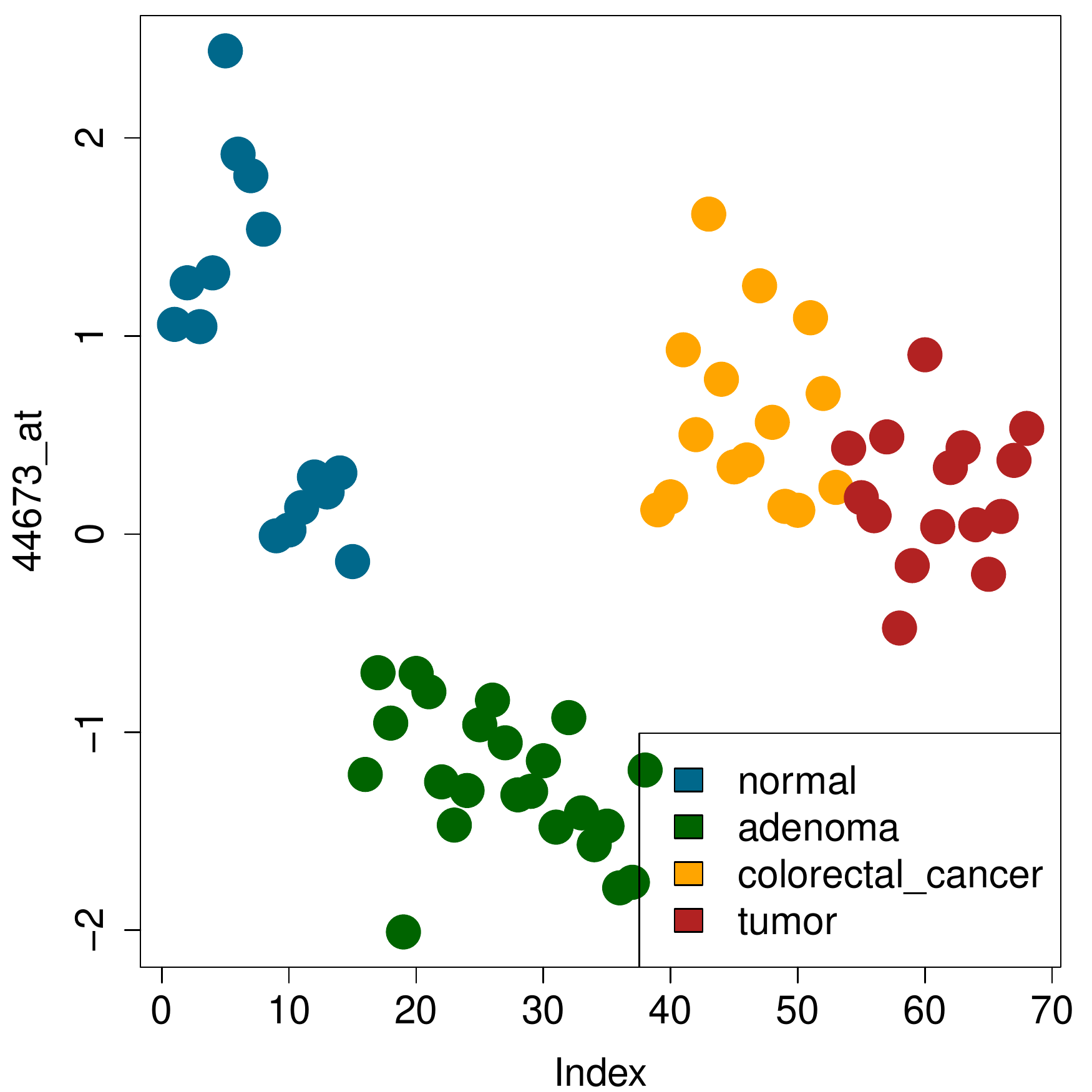}
\includegraphics[width = 0.49\textwidth]{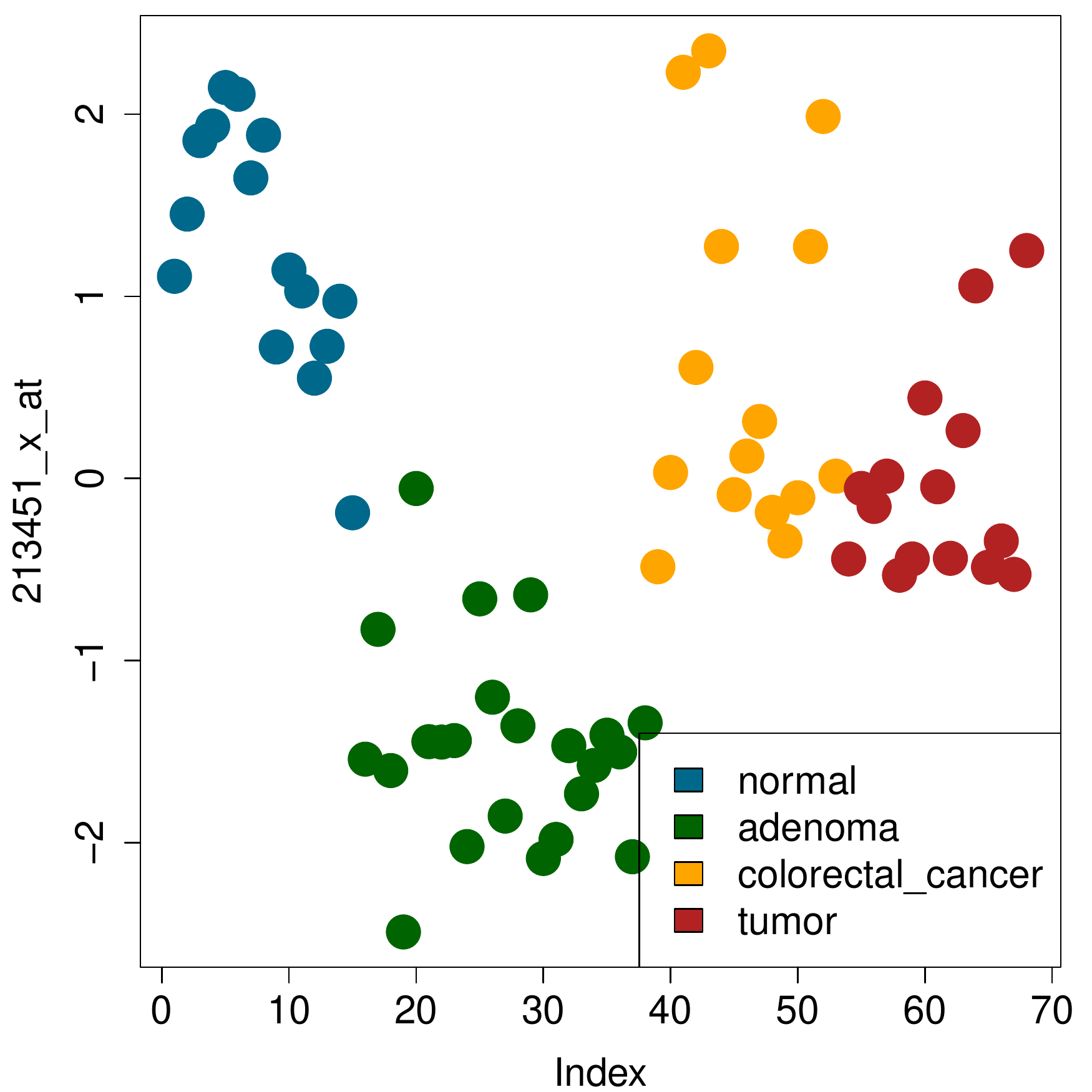}
\caption{Expression levels of the first 4 features entering the clustering model.}
\label{fig:coloncancerFull_firstvariables}
\end{figure}

\clearpage
We now consider the full dataset, with all 4 classes. This clustering task is significantly more difficult, as evidenced by the ARI of 0.61 achieved by classical $K$-means. Figure \ref{fig:coloncancerFull_path} shows the regularization path of HT $K$-means on the left and a plot with the model size and ARI on the right. The regularization path is a little bit more noisy than before, but clearly shows unequal importance of the variables in the clustering. On the right panel of the plot, we see that the ARI increases with $\lambda$, up to roughly 0.68 which is reached by $\lambda$ parameters between 0.56 and 0.84. The corresponding models use between 7 and 638 variables to cluster the data. Given that we can essentially reach the 0.68 ARI with only 7 variables, it is interesting to look at the variables which first enter the model. The first 4 variables are shown in Figure \ref{fig:coloncancerFull_firstvariables}. Interestingly, the variable which first enters the model, named \texttt{204719\_at}, was also among the first variables entering when we only considered the normal and rumor categories. This variable seems very important as it clearly distinguishes between healthy tissue and non-healthy tissue of different types. We further see that the other variables which enter the model early mainly distinguish between the adenoma and the other tissue. They also suggest the existence of a sub-cluster within the normal tissue, as all three of the variables \texttt{1552863\_a\_at}, \texttt{44673\_at} and \texttt{213451\_x\_at} indicate a difference between the first 8 and the last 7 blue dots. It turns out that these observations correspond with tissue collected from the rectum mucosa instead of the colon, and so it can in fact be considered a sub-cluster. Finally, the fact that this classification task is much more difficult is also suggested to be the consequence of the difficult separation of the colorectal cancer and tumor tissue.

\clearpage
\bibliographystyle{Chicago}
\bibliography{TempBib}

\clearpage

\appendix
\section*{Supplementary material}
Supplementary material with proofs and additional simulation results available upon request.

\end{document}